\documentclass{article} % For LaTeX2e
\usepackage{iclr2025_conference,times}

\usepackage{amsmath,amsfonts,bm}

\def\eqref#1{equation~\ref{#1}}
\def\1{\bm{1}}

\DeclareMathAlphabet{\mathsfit}{\encodingdefault}{\sfdefault}{m}{sl}
\SetMathAlphabet{\mathsfit}{bold}{\encodingdefault}{\sfdefault}{bx}{n}

\usepackage{hyperref}
\usepackage{url}
\usepackage{multirow}
\usepackage{graphicx}
\usepackage{wrapfig}

\usepackage[utf8]{inputenc}
\usepackage[T1]{fontenc}
\usepackage{xcolor}
\usepackage[most]{tcolorbox}
\usepackage{varwidth}
\usepackage{listings}

\newcommand{\ul}[1]{\underline{#1}}

\newenvironment{preservelinebreaks}
 {\begingroup\obeylines\begingroup\obeyspaces}
 {\endgroup\endgroup}

\newenvironment{systemmessage}
 {\begin{tcolorbox}[
    enhanced,
    breakable,
    colback=black!5,
    colframe=black,
    title=System,
    fonttitle=\bfseries
 ]\begin{preservelinebreaks}\raggedright}
 {\end{preservelinebreaks}\end{tcolorbox}}

\newenvironment{usermessage}
 {\begin{tcolorbox}[
    enhanced,
    breakable,
    colback=green!5,
    colframe=green!50!black,
    title=User,
    fonttitle=\bfseries
 ]\begin{preservelinebreaks}\raggedright}
 {\end{preservelinebreaks}\end{tcolorbox}}

\newenvironment{assistantmessage}
 {\begin{tcolorbox}[
    enhanced,
    breakable,
    colback=blue!5,
    colframe=blue!50!black,
    title=Assistant,
    fonttitle=\bfseries
 ]\begin{preservelinebreaks}\raggedright}
 {\end{preservelinebreaks}\end{tcolorbox}}

\title{LLMs Can Plan Only If We Tell Them}

\author{Bilgehan Sel \thanks{Corresponding author}\\
Department of ECE\\
Virginia Tech\\
Blacksburg, VA 24060, USA \\
\texttt{bsel@vt.edu}
\AND
Ruoxi Jia \\
Department of ECE\\
Virginia Tech\\
Blacksburg, VA 24060, USA \\
\texttt{ruoxijia@vt.edu} \\
\And
Ming Jin \\
Department of ECE\\
Virginia Tech\\
Blacksburg, VA 24060, USA \\
\texttt{jinming@vt.edu} \\
}

\iclrfinalcopy % Uncomment for camera-ready version, but NOT for submission.
\begin{document}

\maketitle

\begin{abstract}
    Large language models (LLMs) have demonstrated significant capabilities in natural language processing and reasoning, yet their effectiveness in autonomous planning has been under debate. While existing studies have utilized LLMs with external feedback mechanisms or in controlled environments for planning, these approaches often involve substantial computational and development resources due to the requirement for careful design and iterative backprompting. Moreover, even the most advanced LLMs like GPT-4 struggle to match human performance on standard planning benchmarks, such as the Blocksworld, without additional support. This paper investigates whether LLMs can independently generate long-horizon plans that rival human baselines. Our novel enhancements to Algorithm-of-Thoughts (AoT), which we dub AoT+, help achieve state-of-the-art results in planning benchmarks out-competing prior methods and human baselines all autonomously.
\end{abstract}

\section{Introduction}
Large language models (LLMs) based on the transformer architecture \citep{vaswani2017attention} have emerged as a transformative force in artificial intelligence, revolutionizing natural language processing and demonstrating remarkable capabilities across diverse domains. These models, trained on vast corpora of text data, have shown prowess not only in language-related tasks but also in problem-solving \citep{huang2022towards}, reasoning \citep{brown2020language, chowdhery2022palm}, and even coding \citep{chen2021evaluating, thoppilan2022lamda}. The rapid advancements in AI technology have sparked intense interest in exploring their potential for more complex cognitive tasks, reinforcement learning \citep{khattar2023winning, khattar2024cmdp, gu2024balance, gu2025safe, meimand2023tuneopt}, control \citep{sel2021comparative, gunes2023dynamic, coskun2022magnetic, sel2021glsdc, ul2024enhancing, ul2022learning}, optimization \citep{jin2023solution, jin2024democratizing, al2023decision, khattar2024optimization, al2024does, yang2023certifiably}, federated-learning \citep{khan2023pi}, cyber-security \citep{manzoor2024zero, roy2024defense, cody2022discovering, huang2022exposing, wang2024leveraging}, including sequential decision-making and planning. These efforts have yielded promising results, showcasing the models' ability to generate solutions for a wide array of challenges \citep{huang2022towards, suzgunChallengingBIGBenchTasks2022}. However, as the complexity of tasks increases, particularly in domains requiring long-horizon planning and precise execution, the limitations of current LLM-based approaches become apparent \citep{yao2022react, long2023large, valmeekam2023planning, sel2024skin}. 

One of the primary challenges in utilizing LLMs for planning tasks is their inherent difficulty in self-verifying outputs \citep{stechly2024self, liu2024trustworthiness, roy2024beyond}. This limitation manifests in various ways, from suggesting potentially illegal actions to failing to recognize whether a goal has been achieved in planning problems. Moreover, LLMs often struggle with inductive reasoning, finding it challenging to verify whether their hypotheses hold true for given cases – a crucial step in generalizing from simple observations to broader phenomena \citep{qiu2023phenomenal}. To address these shortcomings, researchers have increasingly turned to hybrid approaches that combine LLMs with external verification tools \citep[][\textit{inter alia}]{zhouLeasttoMostPromptingEnables2022, drozdovCompositionalSemanticParsing2022, yao2024tree}. These methods typically use LLMs as idea generators or heuristics, with external systems providing feedback to guide the models towards more accurate and feasible solutions. While this approach has shown promise, it introduces significant complexity and potential failure modes, often requiring substantial development time and resources to implement and maintain \citep{chu2023survey}.

An alternative strand of research has focused on improving LLMs' planning capabilities through advanced prompting techniques. Mirroring biological cognitive abilities to activate analytical \textit{System 2} instead of immediate \textit{System 1} \citep{kahneman2011thinking}, methods such as Chain-of-Thought (CoT) aimed at teaching LLMs step-by-step reasoning, have demonstrated some success in enhancing LLMs' reasoning abilities \citep{wei2022chain, kojima2022large}. However, recent studies have highlighted the limitations of these approaches, particularly in more challenging planning domains. For instance, in benchmarks like Blocksworld, where human performance reaches approximately 78\% accuracy, even state-of-the-art models like GPT-4 achieve only around 30\% accuracy, with CoT prompting offering little to no improvement. The situation is even more concerning in complex planning domains such as logistics, where GPT-4's success rate for generating valid plans drops to a mere 14\% \citep{valmeekam2023planning}. Notably, the performance of open-source LLMs in these planning tasks is even more limited, further underscoring the significant gap between current LLM capabilities and human-level planning proficiency across various domains.

\begin{figure}[t]
    \centering
    \includegraphics[width=0.9\linewidth]{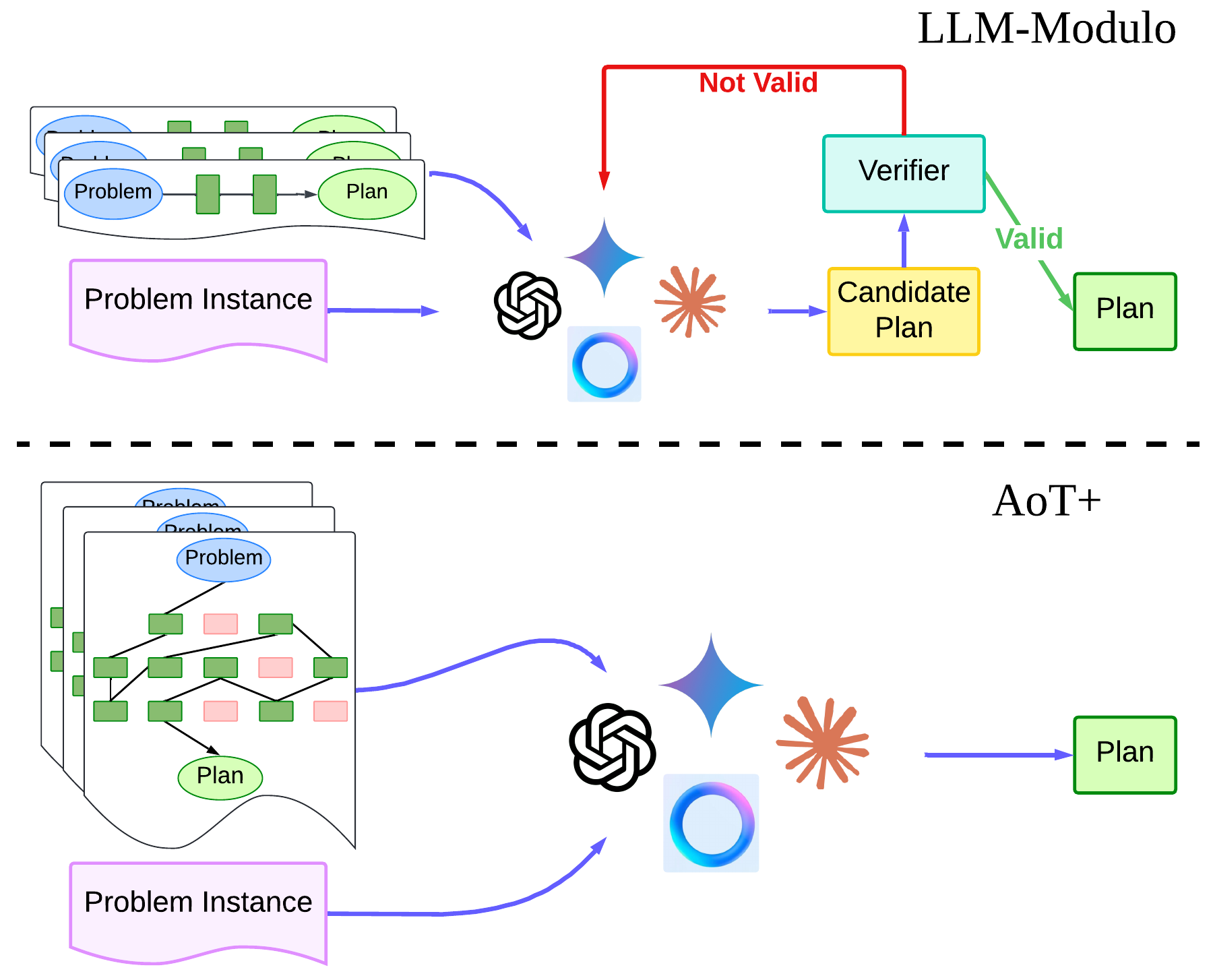}
    \caption{Illustration outlining differences between AoT+ and LLM-Modulo-like frameworks.}
    \label{fig:aot_plus_illustration}
\end{figure}

These sobering results have led some researchers to conclude that current LLMs are fundamentally ill-suited for autonomous planning tasks, particularly those requiring long-horizon reasoning \citep{valmeekam2023planning, stechly2024self, kambhampati2024llms}. However, our in-depth investigations suggest that this pessimism may be premature. We posit that with simple in-context examples showing the search process that also acknowledges the limitations of LLMs, it is possible to dramatically improve their autonomous planning capabilities without resorting to external verification tools which significantly increases development and computational costs.

In this paper, we introduce AoT+, an enhanced prompting technique that builds upon the foundation of the Algorithm of Thoughts (AoT) approach \citep{sel2024algorithm} to activate \textit{System 3} thinking, a more deliberate decision-making process one uses when facing doubt, dilemma, or disruption \citep{webb2021system}. Our method not only challenges the perceived limitations of LLMs in planning tasks but also demonstrates the potential to surpass previous state-of-the-art results, including those achieved using external verifiers. Through extensive experimentation and analysis, we seek to answer two critical questions:
\begin{enumerate}
    \item Can LLMs generate long-horizon plans that rival human performance without external tools?
    \item If so, what key factors differentiate our prompting technique from other step-by-step methods like Chain-of-Thought?
\end{enumerate}

To address these questions, we first examine the limitations of existing approaches. We argue that while CoT prompting has shown success in many reasoning tasks, it falls short in non-ergodic planning problems where a single misstep can lead to an unrecoverable state. Methods like Tree of Thoughts (ToT) \citep{yao2024tree} attempt to mitigate this issue by using external systems to track the search process and guide exploration. However, the computational cost of these approaches renders them impractical for problems with non-trivial depth and breadth. AoT framework \citep{sel2024algorithm}, which demonstrated that incorporating ``human intuitions'' in the search process, along with self-verification and backtracking mechanisms, could lead to significant improvements over CoT while remaining computationally efficient. However, our preliminary tests with vanilla AoT revealed a tendency for state hallucinations, which we hypothesize is caused by excessive ``cognitive load'' – a phenomenon where the model struggles to manage all the relevant information, leading to errors in decision-making.

Motivated by recent findings on LLMs' attention mechanisms, which show a tendency to focus on the beginning and end of their context \citep{liu2024lost}, we introduced several key innovations in AoT+:
\begin{enumerate}
    \item Periodic Structured State Generation: We implement a mechanism for periodically regenerating and restating the current problem state during the search process. This helps alleviate the cognitive load on the model by reducing the need to attend to the entire context history.
    \item Random Trajectory Augmentation: To further simplify the prompting process and improve generalizability, we introduced random search trajectories augmented with correct steps leading to goal states. This approach allows for more efficient example generation without requiring human-authored thought processes.
\end{enumerate}
Figure \ref{fig:aot_plus_illustration} illustrates the key components and workflow of our AoT+ method compared to traditional approaches. These innovations have led to remarkable improvements in performance across multiple challenging planning domains. In the Blocksworld benchmark, AoT+ not only surpasses previous LLM-based methods but also exceeds human performance benchmarks. Similarly, in the logistics domain, our method achieves state-of-the-art results, dramatically improving upon the performance of vanilla GPT-4. The success of AoT+ raises intriguing questions about the nature of reasoning in LLMs and the potential for unlocking more sophisticated cognitive abilities through better understanding. Our work suggests that LLMs may possess latent planning capabilities that can be activated through the right combination of context, structure, and guidance.

\section{Related Work}
\paragraph{Sequential Decision-Making with LLMs.} Having been trained on a large corpus of world-wide text, LLMs excel at understanding a wide range of topics that helps them coming up with possible continuations. The earliest works have observed improvements over standard prompting \citep{brown2020language} for general problem solving, where we directly expect the model to generate the steps one after the other, by step-by-step reasoning by transforming the original problem to a sequential decision-making one, e.g., CoT \citep{nye2021show, wei2022chain, kojima2022large, zhang2022automatic}. However, ToT showed the underwhelming performance in problems that are inherently sequential, such as planning problems \citep{long2023large, yao2024tree}. These works and their follow-ups,  \citet{lei2023boosting, besta2024graph, chen2024boosting}, have relied on using LLMs as mere heuristics with an external mechanism to keep track of search traces to further boost LLMs' capabilities. However, due to significantly increased API requests for individual search nodes, they are notoriously expensive and slow. \citet{sel2024algorithm} proposed the use of pure LLM framework that requires only a single query to match or even surpass methods like ToT, using carefully curated examples woven with human-intuitions in their search trajectories directly in-context, that leads to drastic reductions in compute and API costs.

\paragraph{Self-Verification.} It is an intuitive and natural direction to try to utilize LLMs to evaluate the reasonability and the correctness of their decisions. There is large literature showing that self-verification can help detect mistakes in the token generation of LLMs to correct them to improve performance on domains such as instruction-tuning \citep{bai2022constitutional}, coding \citep{zelikman2023self, kim2024language}, ethical decision-making \citep{ma2023let, sel2024skin}, question-answering \citep{madaan2024self, shinn2024reflexion, paul2024refiner, xie2024self, he2024can}. However, there are also cases where LLMs perform poorly at correcting themselves, especially in symbolic tasks \citep{valmeekam2023planning, kamoi2024can, kambhampati2024llms}.

\paragraph{Classical Search Algorithms.} The field of classical search algorithms has a rich history in AI and planning. Dynamic Programming \citep{bellman1966dynamic}, laid the groundwork for solving complex problems by breaking them down into simpler subproblems. The A* algorithm \citep{hart1968formal}, revolutionized pathfinding and graph traversal by combining the benefits of breadth-first search and best-first search. More recently, Monte Carlo Tree Search (MCTS) methods \citep{kocsis2006bandit, coulom2006efficient, gelly2007combining, ramadan2023monte}, exemplified by AlphaGo \citep{silver2017mastering} and AlphaZero \citep{schrittwieser2020mastering}, have shown remarkable success in game-playing domains, demonstrating the power of combining search with learned policies. The concept of heuristics, central to many of these algorithms, guides the search process towards promising solutions, a principle we leverage in our use of LLMs as heuristic generators.

\section{Prompting Methodologies for Planning Problems}

\subsection{Planning vs. Myopic Problems}
To understand the challenges faced by Large Language Models (LLMs) in planning tasks, it is crucial to distinguish between myopic and planning problems \citep{keeney1993decisions, bertsekas1995dynamic}:

\textbf{Myopic Problems.} A myopic problem is a task that can be solved through simple reasoning and memorization, typically requiring a straightforward, step-by-step approach without the need for long-term strategy or consideration of future consequences.

\textbf{Planning Problems.} A planning problem is a task that requires the ability to formulate a sequence of actions to achieve a specific goal, often involving multiple steps, consideration of future states, and the ability to backtrack or revise the plan based on intermediate outcomes.

The key distinction lies in the cognitive processes required for each type of problem. Myopic problems can often be solved using a predetermined set of steps, making them amenable to simple prompting techniques. Planning problems, however, demand a more sophisticated approach that incorporates: \textit{Self-doubt and verification} (the ability to question and verify each step's validity and its contribution to the overall goal); \textit{Heuristic reasoning} (the use of intuition or learned strategies to guide the exploration of more promising solution paths); \textit{Backtracking} (the capability to recognize dead-ends and return to previous states to explore alternative paths); and \textit{State-tracking} (maintaining an accurate representation of the current problem state throughout the solution process). These requirements pose significant challenges for LLMs, which are primarily trained on static text corpora and may lack explicit training in dynamic problem-solving scenarios. This discrepancy manifests in curious phenomena: LLMs can often generate code/plans to solve planning problems but struggle to execute the same logic in natural language reasoning tasks. We posit that this disconnect stems from the nature of the training data and the inherent limitations of current prompting methodologies.

\subsection{The Incompatibility of Chain-of-Thought in Planning Problems}
Chain-of-Thought (CoT) prompting has emerged as a popular technique for enhancing LLMs' reasoning capabilities. However, current literature reveals fundamental incompatibilities between CoT and the requirements of planning problems:

\begin{enumerate}
    \item \textbf{Linear thinking}: CoT encourages a linear progression of thoughts, which is often insufficient for problems requiring exploration of multiple paths or backtracking \citep{stechly2023gpt, sel2024skin}.
    \item \textbf{Lack of self-correction}: The step-by-step nature of CoT does not inherently support the recognition and correction of mistakes made early in the reasoning process \citep{yao2022react}.
    \item \textbf{Overreliance on example structure}: LLMs tend to mimic the structure of provided examples, leading to rigid thinking patterns that may not generalize well to novel problem instances \cite{sel2024algorithm}.
\end{enumerate}

\begin{figure}
    \centering
    \includegraphics[width=0.99\linewidth]{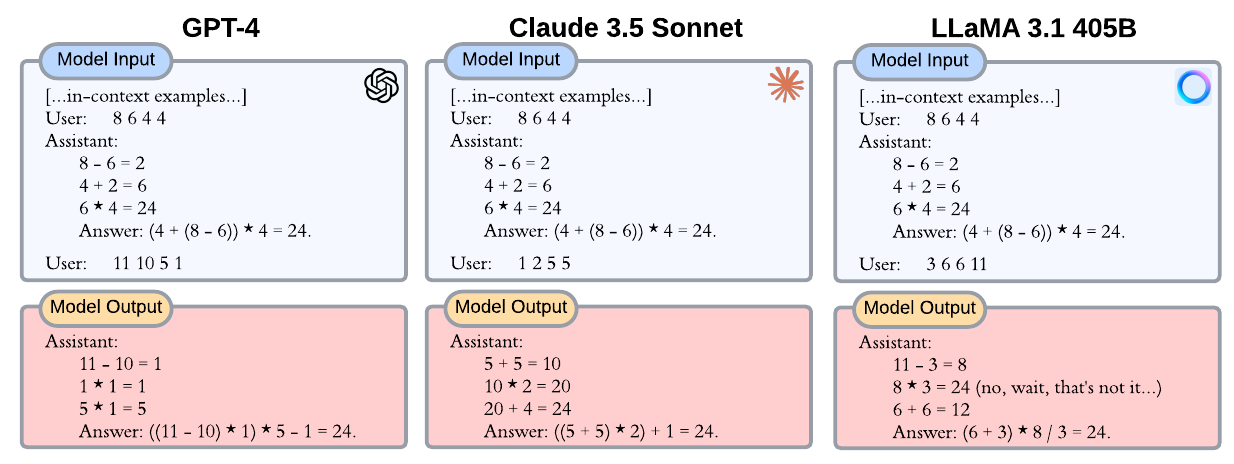}
    \caption{Observed tendency of state-of-the-art LLMs to make simple arithmetic errors when following Chain of Thought (CoT) prompting.}
    \label{fig:incompatibility_of_cot}
\end{figure}

To illustrate these limitations, we conducted experiments using the Game of 24, a simple yet illustrative planning problem with a depth of 3 and a maximum breadth of 48. Figure \ref{fig:incompatibility_of_cot} demonstrates how major LLMs, when presented with CoT examples, tend to produce responses that stylistically match the examples but often fail to arrive at correct solutions.

This observation underscores a critical insight: the effectiveness of prompting techniques can be heavily influenced by the distribution of problem-solving approaches in the training data. The prevalence of step-by-step solutions in educational contexts may inadvertently bias frontier LLMs towards CoT-like reasoning, limiting their ability to adapt to problems requiring more flexible thinking.

While traditional planning algorithms like A* or MCTS can cleanly separate planning from execution, this separation becomes less clear when considering LLM-based planning. In real-world applications where we rely on LLMs, the action space is often vast or infinite, and the execution itself may require complex natural language generation (e.g., creative writing) or reasoning (e.g., crossword puzzles) that cannot be easily reduced to simple programmatic execution. Even in seemingly straightforward domains like Blocksworld, our experiments reveal that LLMs struggle with maintaining accurate state representations during plan execution, as evidenced by the increasing error rates in both CoT and vanilla AoT approaches (see Appendix \ref{appendix:solution_depth}).

\subsection{Algorithm-of-Thoughts Prompting for Planning}
The Algorithm-of-Thoughts (AoT) prompting technique represents a significant advancement in addressing the limitations of CoT for planning problems. Key features of AoT include:

\begin{itemize}
    \item \textbf{Explicit search process}: AoT incorporates a more verbose description of the problem-solving steps, including exploration of multiple paths.
    \item \textbf{Backtracking examples}: In-context examples demonstrate the process of backtracking when reaching dead-ends, teaching LLMs that direct paths to solutions are not always available.
    \item \textbf{Heuristic guidance}: AoT prompts include human-like intuitions to guide the search process, mimicking expert problem-solving strategies.
\end{itemize}

AoT shows marked improvements over CoT in various planning domains, including the Game of 24, crossword puzzles, and creative writing tasks. However, AoT is not without its drawbacks:

\begin{enumerate}
    \item \textbf{Complexity of prompt creation}: The requirement for human-like intuitions in the search process makes crafting effective AoT prompts time-consuming and challenging.
    \item \textbf{Potential for bias}: The inclusion of human heuristics may inadvertently introduce biases or limit the LLM's ability to discover novel solution strategies.
    \item \textbf{State hallucination}: While AoT reduces false positives (invalid solutions), it still struggles with accurately maintaining the problem state throughout the reasoning process.
\end{enumerate}

The issue of state hallucination is particularly intriguing. Our analysis reveals that these hallucinations occur not just at the conclusion of the reasoning process but throughout the solution attempt. This suggests that while AoT improves the overall planning capabilities of LLMs, it does not fully address the fundamental challenge of maintaining an accurate internal representation of the problem state.

These findings motivate our research into more advanced prompting techniques that can better leverage the latent capabilities of LLMs while addressing the specific challenges of planning problems. In the following sections, we introduce our novel AoT+ methodology, which builds upon the strengths of AoT while incorporating mechanisms to mitigate its weaknesses, particularly in the areas of state tracking and heuristic discovery.

\section{AoT+ Prompting}
Motivated by our new understanding of the failure modes in AoT prompting and the challenges in developing prompts that include human-like intuition in the search process, we propose enhancements that drastically improve the performance of LLMs in benchmarks where they were previously shown to be inadequate.

\subsection{Use of Random Solution Traces Does Not Degrade Performance}
While including in-context examples showing the search process improves performance, the requirement for these examples to incorporate human intuition makes development more involved and potentially arbitrary. To support the notion that LLMs can plan autonomously, we tested completely random trajectories, only interwoven with the correct solution path that reaches the goal at the end.

We utilize a novel approach to generate search trajectories by combining successful and unsuccessful solution attempts. Starting with one successful trajectory that reaches the goal state and four unsuccessful ones, we first select a random number of steps from the initial solving process. We then intersperse these with random jumps between states drawn from the unsuccessful trajectories, again selecting a random number of steps at each transition. Crucially, we ensure that our in-context examples always terminate with the final successful steps that reach the goal state from the successful trajectory. This approach introduces controlled randomness while maintaining goal-directed behavior - the random portions allow exploration of the search space, while consistently ending with successful goal achievement creates an implicit bias that helps guide the model toward finding valid solutions. Despite the predominantly random nature of these trajectories, our empirical results show that this method effectively engages the model in active search behavior. The randomness appears to help prevent the model from fixating on specific solution patterns while still maintaining enough structure through the guaranteed successful conclusion to guide it toward valid solutions.

Contrary to our expectations that this approach would significantly increase solution length and the chance of hallucination, we found that random trajectories have a negligible impact on performance. Our comprehensive results across all three benchmarks originally tested in the AoT paper (which are also used in Tree of Thoughts paper) demonstrate this surprising finding.

\begin{table}[h]
\centering
\begin{tabular}{lccc}
\hline
\textbf{Method} & \multicolumn{1}{l}{\textbf{Game of 24}} & \multicolumn{1}{l}{\textbf{Crossword Puzzle}} & \multicolumn{1}{l}{\textbf{Creative Writing}} \\ \hline
CoT-SC          & 9.0\%                                   & 15.6\%                                        & 6.93                                          \\
ToT             & 69.0\%                                  & 46.5\%                                        & 7.56                                          \\
AoT             & \textbf{71.0\%}                                  & 52.0\%                                        & 7.58                                          \\
AoT+R           & 70.0\%                                  & \textbf{54.0\%}                                        & \textbf{7.59}                                          \\ \hline
\end{tabular}
\caption{The effect of utilizing random trajectories instead of human intuition for AoT+ in various benchmarks.}
\label{tab:aot_random}
\end{table}

Table \ref{tab:aot_random} presents a comparison of performance across different methods such as CoT-SC (self-consistency) \citep{wangSelfConsistencyImprovesChain2022} with 10 samples, including Tree of Thoughts (ToT), original AoT, and our random trajectory version of AoT (referred to as AoT+R(andom)). Notably, AoT+R achieves performance very close to that of original AoT, and both surpass ToT across all benchmarks. This is particularly significant given that ToT relies on external tools for state tracking and management. These results suggest that the power of the AoT approach lies not in the specific heuristics provided, but in the overall structure of the problem-solving process it encourages. By demonstrating that random trajectories can be nearly as effective as carefully crafted ones, we open the door to more flexible and generalizable prompting strategies for planning problems.

\subsection{Memoization Avoids Hallucinations}
Our analysis revealed frequent hallucinations in state representation during the AoT process. We hypothesize that these hallucinations stem from the LLM's need to continuously recompute and track the current state after each action, potentially overwhelming its computational capacity as the solution trace grows longer.

\begin{figure}[t]
    \centering
    \includegraphics[width=0.99\linewidth]{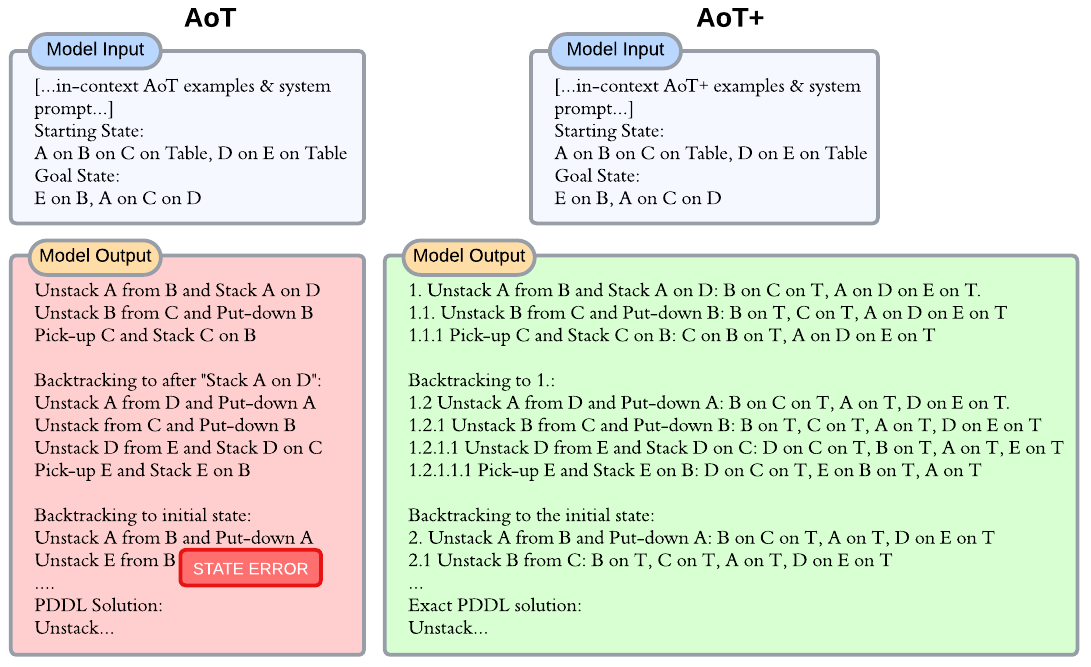}
    \caption{Comparison between AoT and AoT+ for Blocksworld benchmark. Due to AoT's computational overhead of reprocessing the entire context, it hallucinates state and produces an action for another state. AoT+ on the other hand, periodically restates and caches the current problem to hop to any previously visited node.}
    \label{fig:aot_and_aot_plus}
\end{figure}

To address this issue, we draw an analogy to the concept of memoization in dynamic programming. In computer science, memoization is an optimization technique that stores the results of expensive function calls and returns the cached result when the same inputs occur again. We adapt this principle to our prompting strategy, periodically restating and caching the current problem state with identifiers such as ``$x$.$y$.$z$.'' where $x$ is the LLM's $x$-th candidate for the first decision step, and $y$ represents the $y$-th candidate for the second operation after $x$-th candidate for the first one. throughout the solution process as shown in Figure \ref{fig:aot_and_aot_plus}.

This approach offers several advantages over external state tracking methods used in techniques like ToT:
\begin{enumerate}
    \item It eliminates the need for external models to interpret actions and compute states, which can be complex and error-prone.
    \item It avoids the computational overhead of reprocessing the entire context when new information is added, leveraging the caching mechanisms inherent in transformer architectures.
    \item It significantly reduces API costs and latency in real-world applications, as it doesn't require stopping and restarting the generation process to inject external state information.
\end{enumerate}

\begin{wraptable}{r}{0.5\textwidth}
\centering
\begin{tabular}{lll}
\hline
Method & Blocksworld & List Functions \\ \hline
AoT    & 86.3\%      & 78.5\%         \\
AoT+   & 27.0\%      & 26.3\%         \\ \hline
\end{tabular}
\caption{Average of the percentage of the attention over the solution steps (in opposed to the problem definition and/or initial and goal state definitions.}
\label{tab:attention_patterns}
\end{wraptable}

To validate our hypothesis and demonstrate the effectiveness of this memoization-inspired approach, we conducted experiments using LLaMA-2-13B-chat offered by Meta to observe changes in attention patterns. Table \ref{tab:attention_patterns} demonstrates the more structured attention mechanism as a shift towards the visited states, resulting from our AoT+ approach with memoization. This structured attention suggests that the model can more easily access and utilize relevant state information throughout the reasoning process without resulting in too much state identification.

The incorporation of memoization in AoT+ addresses a fundamental limitation in how LLMs process long sequences of information in planning tasks. By providing periodic, easily accessible state summaries, we reduce the cognitive load on the model, allowing it to focus more on the planning process itself rather than struggling to maintain an accurate representation of the problem state.

% \sel{Add a computational theoretical reason why memoization is needed.}

It is worth noting that this approach to state management in prompting shares similarities with the challenges faced in inductive reasoning tasks. Both planning and induction problems require the model to maintain and verify hypotheses over extended reasoning chains. The success of our memoization technique in planning tasks suggests potential applications in improving LLM performance on inductive reasoning problems as well.

\section{Experimental Results}
In this section, we show that our simplified and enhanced prompting version is able to get state-of-the-art results in planning benchmarks, Blocksworld and Logistics, and in inductive reasoning benchmarks, List Functions and ACRE, which are all known to be quite challenging for LLMs \citep{valmeekam2023planning, stechly2024self, qiu2023phenomenal}. We further investigate whether our setups work in a wide range of LLMs.

% \begin{figure}[t]
%     \centering
%     \includegraphics[width=0.9\linewidth]{figures/pipeline.pdf}
%     \caption{An overview of the problem solution pipeline for AoT+.}
%     \label{fig:pipeline}
% \end{figure}

\subsection{Problem Setups}
In this section, we present descriptions of the benchmarks we use, along with prompt generation methodologies for the methods tested. Our problem setups closely follow those in \citet{valmeekam2023planning} for Blocksworld and Logistics, and \citet{qiu2023phenomenal} for ACRE and List Functions. For pure planning problems such as Blocksworld and Logistics, we utilize PDDL to formalize the instances and to check the validity of the outputs. For detailed descriptions of these problem setups, we refer readers to the aforementioned papers.

\textbf{Blocksworld.} Blocksworld is a classic planning domain where the goal is to arrange a set of blocks into a specified configuration. Each block can be on the table or on top of another block, and the agent can perform actions such as picking up a block, putting it down, or stacking it on another block. This domain tests an LLM's ability to reason about spatial relationships and sequential actions.

\textbf{Logistics.} The Logistics domain involves planning the transportation of packages between locations in different cities. It includes trucks for intra-city transport and airplanes for inter-city transport. This domain tests an LLM's ability to reason about complex multi-step plans involving multiple types of actions and objects.

\textbf{List Functions.} The List Functions dataset \citep{rule2020child} evaluates an LLM's ability to induce rules that transform input lists into output lists. These transformations can range from simple operations like selecting specific elements to more complex operations involving multiple steps or conditional logic.

\textbf{ACRE.} The Abstract Causal REasoning (ACRE) dataset \citep{zhang2021acre} tests an LLM's ability to identify causal relationships. It involves determining which objects (referred to as ``Blickets'') can activate a machine based on observed outcomes.

\textbf{Prompt Generation.} For all problems, we generate prompts following the principles outlined in \citet{sel2024algorithm}, with our additional modifications as described in the previous section. Our prompt generation pipeline creates task-specific prompts for various methods:
\begin{itemize}
\item CoT: We provide examples of solved problems with step-by-step reasoning.
\item AoT: We include examples that demonstrate backtracking and exploration of multiple solution paths.
\item AoT+: We incorporate periodic state regeneration along with the random trajectories.
\end{itemize}

For Blocksworld and Logistics, we convert PDDL representations into simple natural language descriptions of the start and goal states. These descriptions serve as the problem instances in our prompts. For List Functions and ACRE, we use natural language to describe the input-output pairs.
In all cases, our prompt generation pipeline allows for flexible creation of task-specific prompts that align with the different methodologies being evaluated, while maintaining consistency with the AoT framework and incorporating our novel enhancements. All the prompts we use for our methods are given in Appendix \ref{appendix:prompts}. For LLM-Modulo frameworks, we use their code-base to evaluate their performance on various LLMs.

\begin{table}[t]
\centering
\begin{tabular}{ccccccc}
\hline
\multirow{3}{*}{\textbf{Problem}} & \multirow{3}{*}{\textbf{Method}} & \multicolumn{5}{c}{\textbf{Accuracy (\%)}}                                                                                                                                                  \\ \cline{3-7} 
                         &                         & GPT-4       & GPT-4o      & Claude  & \begin{tabular}[c]{@{}c@{}}Gemini 1.5\\ 8B-Flash-Pro\end{tabular} & \begin{tabular}[c]{@{}c@{}}LLaMA 3.1\\ 8B-70B-405B\end{tabular} \\ \hline
\multirow{4}{*}{\centering Blocksworld}              & CoT                     & 35          & 34          & 43          & \ul{8}-16-4                                                            & \ul{4}-6-25                                                          \\
                         & LLM-Modulo              & \textbf{82} & \ul{48}    & 19          & 3-4-0                                                             & 0-13-34                                                         \\
                         & AoT                     & 45          & 43          & \ul{66}    & \ul{8}-23-25                                                           & 3-\ul{17}-\ul{35}                                                         \\
                         & AoT+                    & \textbf{82} & \textbf{73} & \textbf{82} & \textbf{27}-\textbf{39}-\textbf{46}                                                          & \textbf{5}-\textbf{52}-\textbf{77}                                                         \\ \hline
\multirow{4}{*}{\centering Logistics}                & CoT                     & 14          & 16          & 27          & 6-7-16                                                            & 2-7-5                                                           \\
                         & LLM-Modulo              & \ul{70}    & \ul{56}    & 26          & 3-5-8                                                             & 4-\ul{16}-30                                                         \\
                         & AoT                     & 24          & 34          & \ul{41}    & \ul{8}-\ul{11}-\ul{24}                                                           & \ul{5}-13-\ul{32}                                                         \\
                         & AoT+                    & \textbf{80} & \textbf{70} & \textbf{53} & \textbf{19}-\textbf{24}-\textbf{57}                                                          & \textbf{14}-\textbf{71}-\textbf{75}                                                        \\ \hline
\multirow{4}{*}{\centering List Functions}           & CoT                     & 38          & 34          & 38          & 18-22-32                                                          & 4-18-28                                                         \\
                         & LLM-Modulo              & \ul{70}    & \ul{66}    & \ul{62}    & \textbf{38}-\ul{54}-\ul{66}                                                          & \ul{18}-34-\ul{54}                                                        \\
                         & AoT                     & 58          & 62          & 44          & 28-32-54                                                          & 14-\ul{36}-42                                                        \\
                         & AoT+                    & \textbf{84} & \textbf{70} & \textbf{64} & \textbf{38}-\textbf{56}-\textbf{68}                                                          & \textbf{28}-\textbf{62}-\textbf{68}                                                        \\ \hline
\multirow{4}{*}{\centering ACRE}                     & CoT                     & 28          & 26          & 22          & 12-18-24                                                          & 8-14-26                                                         \\
                         & LLM-Modulo              & \ul{52}    & 46          & \ul{50}    & \ul{20}-\ul{34}-\ul{52}                                                          & \ul{10}-\ul{34}-\ul{46}                                                        \\
                         & AoT                     & 46          & \ul{48}    & 34          & 18-26-38                                                          & 8-20-42                                                         \\
                         & AoT+                    & \textbf{72} & \textbf{70} & \textbf{56} & \textbf{30}-\textbf{36}-\textbf{58}                                                          & \textbf{20}-\textbf{42}-\textbf{68}                                                        \\ \hline
\end{tabular}
\caption{Performance of various methods on Blocksworld and Logistics environments with various LLMs.}
\label{table:main_results}
\end{table}

\subsection{Main Results}
Our experiments demonstrate the effectiveness of the AoT+ methodology across a range of challenging planning and reasoning tasks. Table \ref{table:main_results} presents a comprehensive comparison of our approach against other methods, including Chain-of-Thought (CoT), LLM-Modulo, and with various LLM architectures. Across all benchmarks---Blocksworld, Logistics, List Functions, and ACRE---AoT+ consistently outperforms or matches the best existing methods, including those using external verification tools like LLM-Modulo. This performance is particularly noteworthy in complex planning domains such as Logistics, where AoT+ shows substantial improvements over both CoT and LLM-Modulo approaches. It also surpasses human performance of 78\% \citep{valmeekam2023planning} in the Blocksworld domain when GPT-4 or Claude is used.

The benefits of AoT+ are evident across different LLM architectures, from GPT-4 to smaller models like LLaMA and Gemini variants. This consistency suggests that our method successfully leverages the inherent capabilities of LLMs, enabling more effective planning and reasoning within a single prompt framework. It is particularly noteworthy that AoT+ consistently outperforms or matches LLM-Modulo across all tasks, despite not relying on external verification tools. This suggests that our method successfully leverages the inherent capabilities of LLMs, enabling them to plan and reason more effectively within a single prompt framework.

The gains of AoT+ are more substantial with larger models, revealing an emergent ability for planning as the scale of the models increases. Notably, the open-source LLaMA 3.1 405B model demonstrates remarkably competitive results with GPT-4 when used with AoT+, a level of performance it fails to achieve within LLM-Modulo frameworks. This observation underscores the effectiveness of AoT+ in unlocking the latent planning capabilities of large language models. The strong performance on both planning (Blocksworld, Logistics) and inductive reasoning (List Functions, ACRE) tasks highlights the versatility of AoT+. By addressing the core challenges of state tracking and exploration in LLM reasoning, our method appears to unlock latent capabilities that are applicable across a wide range of cognitive tasks.

\begin{table}[h]
\centering
\begin{tabular}{ccccccc}
\hline
\multirow{2}{*}{Methods} & \multicolumn{3}{c}{Blocksworld}                    & \multicolumn{3}{c}{Logistics}                       \\ \cline{2-7} 
                         & Input           & Output         & Total           & Input           & Output          & Total           \\ \hline
CoT               & 583.4          & 67.3          & 650.7          & 891.4         & 313.7          & 1205.1         \\
AoT               & 1562.9          & 366.6          & 1929.5          & 2655.3         & 1817.3          & 4472.6         \\
LLM-Modulo               & 5956.0          & 496.9          & 6452.9          & 21201.1         & 1814.2          & 23015.3         \\
AoT+                     & \textbf{1755.2} & \textbf{312.6} & \textbf{2067.8} & \textbf{2879.4} & \textbf{1726.7} & \textbf{5606.1} \\ \hline
\end{tabular}
\caption{Token count comparisons between LLM-Modulo and AoT+ with GPT-4.}
\label{tab:token_count_comparison}
\end{table}

While it might be assumed that AoT+, with its detailed reasoning traces and state-tracking, would require higher total input and output tokens compared to LLM-Modulo, Table \ref{tab:token_count_comparison} reveals a surprising contrast. In fact, AoT+ demonstrates significantly lower token usage, with alternative methods requiring more than 3 times the total tokens. This efficiency is primarily due to us not employing the iterative prompting employed by other frameworks, which rapidly increases both input and output tokens. Moreover, these iterative API requests lead to substantially longer execution times, with LLM-Modulo methods taking on average more than 6 times longer to complete the benchmarks. Although this duration can be influenced by external factors such as internet latency, the magnitude of the difference suggests potential significant impacts in real-time applications, highlighting AoT+'s advantages in scenarios where responsiveness and resource optimization are crucial.

% \section{A\hspace{-0.2em}\textsuperscript{+}oT with External Verifiers}
% \subsection{Simple Backprompting}
% \subsection{Inline Mistake Notification}

% \section{Future Work}
% Systematic exploration of ideas can immensely increase the capabilities of LLMs for sequential decision-making problems. We believe it possible to generate a large dataset on similar problems that require planning to reach a solution, and use it to further instruction-finetune LLMs for them to directly use AoT+ to give responses without any in-context examples.

\section{Conclusion}
This paper introduces AoT+, an enhanced prompting technique that significantly improves the planning and reasoning capabilities of large language models (LLMs). The key innovations of AoT+ address fundamental limitations in how LLMs process long sequences of information in planning tasks. Through comprehensive experiments across challenging benchmarks, our results consistently show that AoT+ matches or outperforms existing SOTA methods, including those using external verification, across various LLM architectures. By demonstrating that LLMs can autonomously plan and reason at high levels of performance, AoT+ opens new avenues for research and applications.

\clearpage
\bibliography{iclr2025_conference}
\bibliographystyle{iclr2025_conference}

\clearpage
\appendix
\section{Additional Experiments}
\subsection{Impact Of Solution Depth for AoT and AoT+}\label{appendix:solution_depth}
In order to provide further evidence to the use of memoization for reducing state errors and hallucinations, we conducted an experiment to analyze the error rates for states for AoT and AoT+ in the Logistics benchmark using LLaMA 3.1 70B model. This model is chosen since it is a relatively cheap to do inference on computationally while already having a good performance in the benchmark with AoT+.

\begin{figure}[h]
    \centering
    \includegraphics[width=0.99\linewidth]{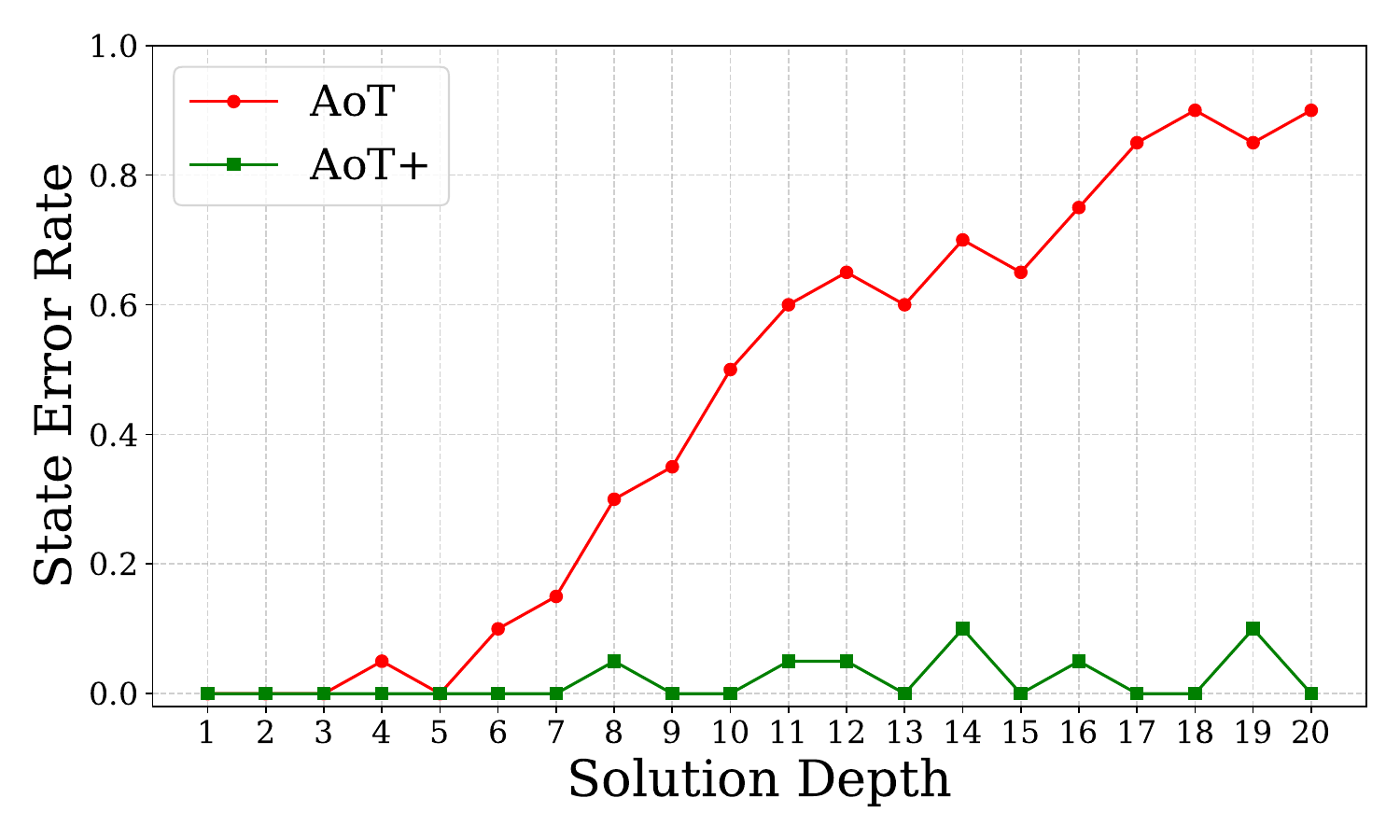}
    \caption{Comparison of error rates in state estimation with respect to solution depth for AoT and AoT+ in the Logistics benchmark using LLaMA 3.1 70B model.}
    \label{fig:depth_impact}
\end{figure}

We chose 200 games from the Logistics benchmark where both AoT+ and AoT was providing solutions after reaching a solution depth of 20 actions, whether it be correct or not. Then we sampled 20 states in each depth and checked whether the state assumed by the LLM would be reached if we were to follow the actions it proposed starting from the initial state. If there was a discrepancy, we marked it as an error. In Figure \ref{fig:depth_impact}, we see that AoT+ dramatically reduces state hallucinations and errors compared to AoT, which in return helps it achieve a superior performance as shown in Table \ref{table:main_results} across various benchmarks and LLMs.

\clearpage
\subsection{Impact of Each Innovation of AoT+}
We also provide a more complete main results together with ablation studies on the impact of each innovation of AoT+. We denoted AoT with random solution traces instead of human intuitions as AoT+R and AoT with memoization as AoT+M. As we can see in Table \ref{table:main_results_full}, AoT+R do have very close performance to AoT, whereas AoT+M, or we can think of it AoT+ with human intuitions, getting similar performance to AoT+.

\begin{table}[h]
\centering
\begin{tabular}{ccccccc}
\hline
\multirow{3}{*}{\textbf{Problem}} & \multirow{3}{*}{\textbf{Method}} & \multicolumn{5}{c}{\textbf{Accuracy (\%)}}                                                                                                                                                  \\ \cline{3-7} 
                         &                         & GPT-4       & GPT-4o      & Claude  & \begin{tabular}[c]{@{}c@{}}Gemini 1.5\\ 8B-Flash-Pro\end{tabular} & \begin{tabular}[c]{@{}c@{}}LLaMA 3.1\\ 8B-70B-405B\end{tabular} \\ \hline
\multirow{4}{*}{\centering Blocksworld}              & CoT                     & 35          & 34          & 43          & \ul{8}-16-4                                                            & \ul{4}-6-25                                                          \\
                         & Self-Refine (SF = 10)              & 39 & 43    & 48          & 12-13-4                                                             & 3-8-25                                                         \\
                         & Tree-Planner (N = 25)              & 44 & 47    & 46          & 11-20-31                                                             & 7-16-33                                                         \\
                         & LLM-Modulo              & \textbf{82} & \ul{48}    & 19          & 3-4-0                                                             & 0-13-34                                                         \\
                         & AoT                     & 45          & 43          & \ul{66}    & \ul{8}-23-25                                                           & 3-\ul{17}-\ul{35}                                                         \\
                         & AoT+R                     & 44          & 44    & 67          & 7-22-26                                                          & 4-16-35                                                        \\
                         & AoT+M                     & 84          & 73    & 84          & 28-39-48                                                          & 6-52-80                                                         \\
                         & AoT+                    & \textbf{82} & \textbf{73} & \textbf{82} & \textbf{27}-\textbf{39}-\textbf{46}                                                          & \textbf{5}-\textbf{52}-\textbf{77}                                                         \\ \hline
\multirow{4}{*}{\centering Logistics}                & CoT                     & 14          & 16          & 27          & 6-7-16                                                            & 2-7-5                                                           \\
                         & LLM-Modulo              & \ul{70}    & \ul{56}    & 26          & 3-5-8                                                             & 4-\ul{16}-30                                                         \\
                         & AoT                     & 24          & 34          & \ul{41}    & \ul{8}-\ul{11}-\ul{24}                                                           & \ul{5}-13-\ul{32}                                                         \\
                         & AoT+R                     & 24          & 35   & 41          & 9-10-24                                                          & 7-16-36                                                         \\
                         & AoT+M                     & 80          & 72    & 51          & 18-25-58                                                          & 15-73-77                                                         \\
                         & AoT+                    & \textbf{80} & \textbf{70} & \textbf{53} & \textbf{19}-\textbf{24}-\textbf{57}                                                          & \textbf{14}-\textbf{71}-\textbf{75}                                                        \\ \hline
\multirow{4}{*}{\centering List Functions}           & CoT                     & 38          & 34          & 38          & 18-22-32                                                          & 4-18-28                                                         \\
                         & LLM-Modulo              & \ul{70}    & \ul{66}    & \ul{62}    & \textbf{38}-\ul{54}-\ul{66}                                                          & \ul{18}-34-\ul{54}                                                        \\
                         & AoT                     & 58          & 62          & 44          & 28-32-54                                                          & 14-\ul{36}-42                                                        \\
                         & AoT+R                     & 56          & 62    & 46          & 28-30-54                                                          & 12-36-42                                                         \\
                         & AoT+M                     & 84          & 72    & 62          & 42-56-66                                                         & 28-62-66                                                         \\
                         & AoT+                    & \textbf{84} & \textbf{70} & \textbf{64} & \textbf{38}-\textbf{56}-\textbf{68}                                                          & \textbf{28}-\textbf{62}-\textbf{68}                                                        \\ \hline
\multirow{4}{*}{\centering ACRE}                     & CoT                     & 28          & 26          & 22          & 12-18-24                                                          & 8-14-26                                                         \\
                         & LLM-Modulo              & \ul{52}    & 46          & \ul{50}    & \ul{20}-\ul{34}-\ul{52}                                                          & \ul{10}-\ul{34}-\ul{46}                                                        \\
                         & AoT                     & 46          & \ul{48}    & 34          & 18-26-38                                                          & 8-20-42                                                         \\
                         & AoT+R                     & 46          & 50    & 34          & 18-28-34                                                          & 6-22-42                                                         \\
                         & AoT+M                     & 70          & 70    & 60          & 36-36-60                                                          & 24-44-58                                                         \\
                         & AoT+                    & \textbf{72} & \textbf{70} & \textbf{56} & \textbf{30}-\textbf{36}-\textbf{58}                                                          & \textbf{20}-\textbf{42}-\textbf{68}                                                        \\ \hline
\end{tabular}
\caption{Performance of various methods on Blocksworld, Logistics, List Functions and ACRE environments with various LLMs. SF refers to the number of iterations for self-feedback, and N refers to the number of initial samples for Tree-Planner.}
\label{table:main_results_full}
\end{table}

\subsection{More Information on Token Efficiency}
\begin{table}[h]
\centering
\begin{tabular}{ccccccc}
\hline
\multirow{2}{*}{Methods} & \multicolumn{3}{c}{Blocksworld}                    & \multicolumn{3}{c}{Logistics}                       \\ \cline{2-7} 
                         & Input           & Output         & Total           & Input           & Output          & Total           \\ \hline
CoT               & 583.4 $\pm 22.6$          & 67.3 $\pm 25.2$          & 650.7 $\pm 33.8$           & 891.4 $\pm 46.6$        & 313.7 $\pm 19.8$      & 1205.1 $\pm 50.6$     \\
AoT               & 1562.9 $\pm 15.3$          & 366.6 $\pm 88.7$          & 1929.5 $\pm 90.0$       & 2655.3 $\pm 174.3$     & 1817.3 $\pm 263.4$    & 4472.6 $\pm 315.8$   \\
LLM-Modulo               & 5956.0 $\pm 37.8$          & 496.9 $\pm 44.4$          & 6452.9 $\pm 58.3$     & 21201.1 $\pm 42.2$   & 1814.2 $\pm 306.7$          & 23015.3 $\pm 309.6$     \\
AoT+                     & 1755.2  $\pm 15.3$ & 312.6 $\pm 105.8$ & 2067.8 $\pm 106.9$ & 2879.4 $\pm 174.3$ & 1726.7 $\pm 208.2$ & 5606.1 $\pm 271.5$  \\ \hline
\end{tabular}
\caption{Token count comparisons between LLM-Modulo and AoT+ with GPT-4 with std also given.}
\label{tab:token_count_comparison_with_std}
\end{table}

\subsection{Additional Baseline - Self-Refinement}
To demonstrate simple iterative methods for improving planning performance through LLM self-feedback, we evaluated Self-Refine \citep{madaan2024self}. We adhered to their original hyperparameters (Temperature = 0.7) but extended the maximum iterations from 4 to 10 to ensure fair comparison with our other baseline, Tree-Planner \citep{hu2023tree}, which employs a maximum of 10 corrections (SF = 10). All prompts for initial generation and feedback are provided in Appendix \ref{appendix:prompts}. For prompt generations, we implemented a 5-shot setting, and specifically for feedback, we leveraged the feedback provided to the LLM in the LLM-Modulo framework, which offers reliable validation through an external validator. As shown in Table \ref{table:main_results_full}, Self-Refine demonstrates only marginal improvements over the CoT baseline, a finding that aligns with the ``Mathematical Reasoning'' task results reported in \citet{madaan2024self}.

\subsection{Additional Baseline - Tree-Planner}
We evaluated Tree-Planner \citep{hu2023tree} on the Blocksworld benchmark, incorporating sampling, merging, and backtracking steps for comparative analysis. It is crucial to note that Tree-Planner requires knowledge of action reversibility, which may be challenging or infeasible in non-ergodic environments. Fortunately, Blocksworld allows straightforward action reversal (e.g., ``Pick-up'' reverses ``Put-down'', ``Unstack'' reverses ``Stack''). We maintained the structural hyperparameters from the original implementation, with $N = 25$ initial samples and a maximum of 10 error corrections. For detailed information about these hyperparameters and the method, readers should consult the original paper. As demonstrated in Table \ref{table:main_results_full}, Tree-Planner consistently outperforms Self-Refine and occasionally matches AoT's performance. While it appears to surpass LLM-Modulo, we attribute this partially to LLM-Modulo's optimization for GPT-4, their primary LLM. For generation prompts, we utilized the Self-Refine Initial Generation prompt detailed in Appendix \ref{appendix:self-refine-blocksworld-initial-generation}. The supplementary material includes implementation code for Tree-Planner in the Blocksworld environment.

\clearpage
\section{All Used Prompts}\label{appendix:prompts}
\subsection{Self-Refine Blocksworld - Initial Generation}\label{appendix:self-refine-blocksworld-initial-generation}

\begin{systemmessage}
I am playing with a set of blocks where I need to arrange the blocks into stacks. Here are the actions I can do

Pick up a block
Unstack a block from on top of another block
Put down a block
Stack a block on top of another block

I have the following restrictions on my actions:
I can only pick up or unstack one block at a time.
I can only pick up or unstack a block if my hand is empty.
I can only pick up a block if the block is on the table and the block is clear. A block is clear if the block has no other blocks on top of it and if the block is not picked up.
I can only unstack a block from on top of another block if the block I am unstacking was really on top of the other block.
I can only unstack a block from on top of another block if the block I am unstacking is clear.
Once I pick up or unstack a block, I am holding the block.
I can only put down a block that I am holding.
I can only stack a block on top of another block if I am holding the block being stacked.
I can only stack a block on top of another block if the block onto which I am stacking the block is clear.
Once I put down or stack a block, my hand becomes empty.
Once you stack a block on top of a second block, the second block is no longer clear.
\end{systemmessage}

\begin{usermessage}
As initial conditions I have that, the Block B is clear, the Block D is clear, the hand is empty, the Block C is on top of the Block A, the Block D is on top of the Block C, the Block A is on the table and the Block B is on the table.
My goal is to have that the Block B is on top of the Block C and the Block D is on top of the Block A.
\end{usermessage}

\begin{assistantmessage}
[PLAN]
unstack the Block B from on top of the Block D
put down the Block B on the table
unstack the Block A from on top of the Block C
stack the Block A on top of the Block C
pick up the Block B from the table
stack the Block B on top of the Block A
[PLAN END]
\end{assistantmessage}

\begin{usermessage}
As initial conditions I have that, the Block A is clear, the hand is empty, the Block A is on top of the Block D, the Block B is on top of the Block C, the Block D is on top of the Block B and the Block C is on the table.
My goal is to have that the Block A is on top of the Block C.
\end{usermessage}

\begin{assistantmessage}
[PLAN]
unstack the Block A from on top of the Block D
put down the Block A
unstack the Block D from on top of the Block B
put down the Block D
unstack the Block B from on top of the Block C
put down the Block B
pick up the Block A
stack the Block A on top of the Block C
[PLAN END]
\end{assistantmessage}

\begin{usermessage}
As initial conditions I have that, the Block B is clear, the Block D is clear, the hand is empty, the Block C is on top of the Block A, the Block D is on top of the Block C, the Block A is on the table and the Block B is on the table.
My goal is to have that the Block B is on top of the Block C and the Block D is on top of the Block A
\end{usermessage}

\begin{assistantmessage}
[PLAN]
unstack Block D from on top of the Block C
put down the block D on the table
unstack Block C from on top of the Block A
stack the Block C on top of the Block B
pick up the Block D from the table
stack the Block D on top of the Block A
unstack the Block C from on top the Block B
stack the Block C on top the Block D
pick up the Block B from the table
stack the Block B on top of the Block C
[PLAN END]
\end{assistantmessage}

\begin{usermessage}
As initial conditions I have that, the Block A is clear, the Block B is clear, the hand is empty, the Block A is on top of the Block C, the Block B is on top of the Block D, the Block C is on the table and the Block D is on the table.
My goal is to have that the Block A is on top of the Block C and the Block B is on top of the Block A.
\end{usermessage}

\begin{assistantmessage}
[PLAN]
unstack the Block B from on top of the Block D
put down the Block B on the table
unstack the Block A from on top of the Block C
stack the Block A on top of the Block C
pick up the Block B from the table
stack the Block B on top of the Block A
[PLAN END]
\end{assistantmessage}

\subsection{Self-Refine Blocksworld - Feedback}
\begin{systemmessage}
I am playing with a set of blocks where I need to arrange the blocks into stacks. Here are the actions I can do

Pick up a block
Unstack a block from on top of another block
Put down a block
Stack a block on top of another block

I have the following restrictions on my actions:
I can only pick up or unstack one block at a time.
I can only pick up or unstack a block if my hand is empty.
I can only pick up a block if the block is on the table and the block is clear. A block is clear if the block has no other blocks on top of it and if the block is not picked up.
I can only unstack a block from on top of another block if the block I am unstacking was really on top of the other block.
I can only unstack a block from on top of another block if the block I am unstacking is clear.
Once I pick up or unstack a block, I am holding the block.
I can only put down a block that I am holding.
I can only stack a block on top of another block if I am holding the block being stacked.
I can only stack a block on top of another block if the block onto which I am stacking the block is clear.
Once I put down or stack a block, my hand becomes empty.
Once you stack a block on top of a second block, the second block is no longer clear.
\end{systemmessage}

\begin{usermessage}
As initial conditions I have that, the Block B is clear, the hand is empty, the Block B is on top of the Block C, the Block C is on top of the Block D, the Block D is on top of the Block A and the Block A is on the table.
My goal is to have that the Block A is on top of the Block C and the Block D is on top of the Block A.

My plan is as follows:

[PLAN]
unstack the Block B from on top of the Block C
put down the Block B
unstack the Block C from on top of the Block D
put down the Block C
unstack the Block D from on top of the Block A
stack the Block D on top of the Block B
pick up the Block A
stack the Block A on top of the Block C
unstack the Block D from on top of the Block B
stack the Block D on top of the Block A
[PLAN END]

Check whether this plan is valid or not:
\end{usermessage}
\begin{assistantmessage}
[CHECK]
The above plan is invalid. The following action at step 1 has an unmet precondition:
unstack the Block A from on top of the Block C
The unmet precondition is:
the Block A is clear
[CHECK]
\end{assistantmessage}

\begin{usermessage}
As initial conditions I have that, the Block D is clear, the hand is empty, the Block A is on top of the Block C, the Block C is on top of the Block B, the Block D is on top of the Block A and the Block B is on the table.
My goal is to have that the Block A is on top of the Block D and the Block D is on top of the Block B.

My plan is as follows:

[PLAN]
unstack the Block D from on top of the Block A
put down the Block D
unstack the Block A from on top of the Block C
put down the Block A
unstack the Block C from on top of the Block B
put down the Block C
pick up the Block D
stack the Block D on top of the Block B
pick up the Block A
stack the Block A on top of the Block D
[PLAN END]

Check whether this plan is valid or not:
\end{usermessage}
\begin{assistantmessage}
[CHECK]
The above plan is valid.
[CHECK]
\end{assistantmessage}

\begin{usermessage}
As initial conditions I have that, the Block D is clear, the hand is empty, the Block A is on top of the Block C, the Block C is on top of the Block B, the Block D is on top of the Block A and the Block B is on the table.
My goal is to have that the Block A is on top of the Block B and the Block D is on top of the Block C.

My plan is as follows:

[PLAN]
unstack the Block D from the Block A
put down the Block D
unstack the Block A from the Block C
put down the Block A
pick up the Block D
stack the Block D on top of the Block C
pick up the Block A
stack the Block A on top of the Block B
[PLAN END]

Check whether this plan is valid or not:
\end{usermessage}
\begin{assistantmessage}
[CHECK]
The above plan is invalid. The following action at step 8 has an unmet precondition:
stack the Block A on top of the Block B
The unmet precondition is:
the Block B is clear
[CHECK]
\end{assistantmessage}

\begin{usermessage}
As initial conditions I have that, the Block A is clear, the hand is empty, the Block A is on top of the Block D, the Block B is on top of the Block C, the Block D is on top of the Block B and the Block C is on the table.
My goal is to have that the Block A is on top of the Block C.

My plan is as follows:

[PLAN]
unstack the Block A from on top of the Block D
put down the Block A
unstack the Block D from on top of the Block B
put down the Block D
unstack the Block B from on top of the Block C
put down the Block B
pick up the Block A
stack the Block A on top of the Block C
[PLAN END]

Check whether this plan is valid or not:
\end{usermessage}
\begin{assistantmessage}
[CHECK]
The above plan is valid.
[CHECK]
\end{assistantmessage}

\subsection{AoT+ Blocksworld}

\begin{systemmessage}
I am playing with a set of blocks where I need to arrange the blocks into stacks. Here are the actions I can do  Pick up a block Unstack a block from on top of another block Put down a block Stack a block on top of another block  I have the following restrictions on my actions: I can only pick up or unstack one block at a time. I can only pick up or unstack a block if my hand is empty. I can only pick up a block if the block is on the table and the block is clear. A block is clear if the block has no other blocks on top of it and if the block is not picked up. I can only unstack a block from on top of another block if the block I am unstacking was really on top of the other block. I can only unstack a block from on top of another block if the block I am unstacking is clear. Once I pick up or unstack a block , I am holding the block. I can only put down a block that I am holding. I can only stack a block on top of another block if I am holding the block being stacked. I can only stack a block on top of another block if the block onto which I am stacking the block is clear. Once I put down or stack a block , my hand becomes empty. I cannot stack a block under another block. I can only stack a block on top of a block. Also, in the last part, do not use a numbered list or something to return the actions, just return them like in the examples. Use 'The PDDL version: bla bla' to return the PDDL code, and do not add anything after the actions.
\end{systemmessage}

\begin{usermessage}
Start: D on A on C on B on table.
Goal: A on D on B.
\end{usermessage}

\begin{assistantmessage}
Firstly, we should satisfy D being on B without B being on A. Since B is on the table, we should remove every block on top of B to later stack first D then A on top of B.

1. Unstack D from A.
2. Put-down D
- D on table, A on C on B on table.
3. Unstack A from C
4. Put-down A
- A on table, D on table, C on B on table.
5. Unstack C from B
6. Put-down C
- A on table, B on table, C on table, D on table.

Now everything is on the table and we need to simply one by one stack the blocks in sequence. So, first D on B, then A on D.

7. Pick-up D
8. Stack D on B
- A on table, C on table, D on B on table.
9. Pick-up A
10. Stack A on D

The PDDL version:
(unstack D A)
(put-down D)
(unstack A C)
(put-down A)
(unstack C B)
(put-down C)
(pick-up D)
(stack D B)
(pick-up A)
(stack A D)
\end{assistantmessage}

\begin{usermessage}
Start: A on table, B on C on table, D on table.
Goal: A on B on D on C.
\end{usermessage}

\begin{assistantmessage}
For the goal state, all blocks need to be stacked on top of each other with C being on the table (because there are only four blocks and all four blocks are on top of each other). Then in sequence, first D, then B, then A needs to be stacked.

1. Pick-up D
2. Stack D on C. C is not clear. Backtracking...

1. Unstack B from C
2. Put-down B
- A on table, B on table, C on table, D on table.

Now that all blocks are on the table, we just have to follow the initial plan of ours.

3. Pick-up D
4. Stack D on C
- A on table, B on table, D on C on table.
5. Pick-up B
6. Stack B on D
- A on table, B on D on C on table.
7. Pick-up A
8. Stack A on B.
- A on B on D on C on table.

The PDDL version:
(unstack B C)
(put-down B)
(pick-up D)
(stack D C)
(pick-up B)
(stack B D)
(pick-up A)
(stack A B)
\end{assistantmessage}

\begin{usermessage}
Start: B on C on A on table, D on table.
Goal: B on D, C on A.
\end{usermessage}

\begin{assistantmessage}
B is clear (clear means no block in on top of it) and it only needs to be put on D, and D is clear also. Then, C should be on A, and it's satisfied currently. Let's start with placing B on D and let's examine again then.

1. Unstack B from C
2. Stack B on D
- B on D on table, C on A on table.

Goal is satisfied.

The PDDL version:
(unstack B C)
(stack B D)
\end{assistantmessage}

\begin{usermessage}
Start: B on table, C on A on D on table.
Goal: B on A on D on C.
\end{usermessage}

\begin{assistantmessage}
For the goal state, all blocks need to be stacked on top of each other with C being on the table (because there are only four blocks and all four blocks are on top of each other). Then in sequence, first D, then A, then B needs to be stacked.

1. Unstack A from D. However, A is not clear, there C on top of it. Backtracking...
1. Unstack C from A
2. Put-down C
- B on table, C on table, A on D on table.

Now that C is on the table, we can unstack A from D.

3. Unstack A from D
4. Put-down A
- A on table, B on table, C on table, D on table.

Now that all blocks are on the table, we just have to follow the initial plan of ours.

5. Pick-up D
6. Stack D on C
- A on table, B on table, D on C on table.
7. Pick-up A
8. Stack A on D
- A on D on C on table, B on table.
9. Pick-up B
10. Stack B on A.
- B on A on D on C on table.

The PDDL version:
(unstack C A)
(put-down C)
(unstack A D)
(put-down A)
(pick-up D)
(stack D C)
(pick-up A)
(stack A D)
(pick-up B)
(stack B A)
\end{assistantmessage}

\begin{usermessage}
Start: D on A on B on C on table.
Goal: D on B, A on C.
\end{usermessage}

\begin{assistantmessage}
D is already on B, but B is not clear. A is on C, but C is not clear. Let's start with making B clear.

1. Unstack D from A
2. Put-down D
- D on table, A on B on C on table.
3. Unstack A from B
4. Put-down A
- A on table, B on C on table, D on table.

Now, we can place D on B to make it D on B on C, which would satisfy one of the goals (D on B). A on C is not satisfied since B is on C. Then B needs to also be unstacked.
5. Unstack B from C
6. Put-down B
- A on table, B on table, C on table, D on table.

Now that all blocks are on the table, we just have to follow the initial plan of ours.

7. Pick-up D
8. Stack D on B
- A on table, D on B on table, C on table.
9. Pick-up A
10. Stack A on C.
- A on C on table, D on B on table.

The PDDL version:
(unstack D A)
(put-down D)
(unstack A B)
(put-down A)
(unstack B C)
(put-down B)
(pick-up D)
(stack D B)
(pick-up A)
(stack A C)
\end{assistantmessage}

\begin{usermessage}
Start: D on B on C on A on table.
Goal: D on C on A on B.
\end{usermessage}

\begin{assistantmessage}
For the goal state, all blocks need to be stacked on top of each other with B being on the table (because there are only four blocks and all four blocks are on top of each other). Then in sequence, first A, then C, then D needs to be stacked.

1. Unstack D from B
2. Put-down D
- D on table, B on C on A on table.
3. Unstack B from C
4. Put-down B
- B on table, C on A on table, D on table.
5. Unstack C from A
6. Put-down C
- A on table, B on table, C on table, D on table.

Now that all blocks are on the table, we just have to follow the initial plan of ours.

7. Pick-up A
8. Stack A on B
- A on B on table, C on table, D on table.
9. Pick-up C
10. Stack C on A
- C on A on B on table, D on table.
11. Pick-up D
12. Stack D on C.
- D on C on A on B on table.

The PDDL version:
(unstack D B)
(put-down D)
(unstack B C)
(put-down B)
(unstack C A)
(put-down C)
(pick-up A)
(stack A B)
(pick-up C)
(stack C A)
(pick-up D)
(stack D C)
\end{assistantmessage}

\subsection{AoT+ Logistics}

\begin{systemmessage}
We are playing a game called logistics where there are multiple cities each city having possibly multiple locations and one of them being an airport. For transit, we have trucks and airplanes. Trucks can go to any location inside the city, and the airplanes can go to any location as long as it's another airport. So basically, trucks are used for inter-city transport whereas the airplanes are between the cities. Our goal is given a package at a specific location, we wish to transport it to the goal location. The naming scheme for the locations are Lx-y, where x and y are both digits, and x designates the city the location is at, y is like its index. Lx-0 are always an airport. Always give the final answer as in the examples -> 'The PDDL version: bla bla' and do not add anything after the steps.
\end{systemmessage}

\begin{usermessage}
Start:
Airplanes: A0 at L1-0, A1 at L1-0
Trucks: T0 at L0-1, T1 at L1-0, T2 at L2-1
Packages: P0 at L2-0, P1 at L2-1, P2 at L1-1, P3 at L0-0, P4 at L2-0, P5 at L2-0

Goal: P0 at L2-0, P1 at L0-1, P2 at L1-0, P3 at L2-0, P4 at L1-0, P5 at L1-1
\end{usermessage}

\begin{assistantmessage}
There are three cities since only locations talked about are L0, L1 and L2. We have two airplanes A0 and A1, and are both in L1-0. Let's first examine which packages need to moved to other cities.

P0: L2 -> L2 (no)
P1: L2 -> L0 (yes)
P2: L1 -> L1 (no)
P3: L0 -> L2 (yes)
P4: L2 -> L1 (yes)
P5: L2 -> L1 (yes)

The packages P0 and P2 only need to moved within the city, and P1, P3, P4 and P5 will need to be transferred via planes to other cities. Let's sort out the within city ones first. P0 needs to be moved from L2-0 to L2-0, so no moves are necessary there. P2 should be moved from L1-1 to L1-0. The truck at C1 is T1 at L1-0, so it needs to go to L1-1 first and go to L1-0 after loading the package.

1. Drive-truck T1 at L1-0 to L1-1 in C1
2. Load-truck P2 to T1 at L1-1
3. Drive-truck T1 at L1-1 to L1-0 in C1
4. Unload-truck P2 from T1 at L1-0

Airplanes: A0 at L1-0, A1 at L1-0
Trucks: T0 at L0-1, T1 at (driven but not change) L1-0, T2 at L2-1
Packages: P1 at L2-1, P3 at L0-0, P4 at L2-0, P5 at L2-0

Now, we are finished with inter-city transports. and are left with the packages P1, P3, P4 and P5. Let's transfer P1 first. P1 needs to go from L2-1 to L0-1. Both locations are not airports so we need to first use a truck to carry it from L2-1 to L2-0 and then airplane to take it to L0-0, then finally another truck to take it from L0-0 to L0-1. The truck in L2 is T2 and it is in L2-1. So we can directly load it to take it to L2-0. The airplanes are in L1-0, so one of them (let's say A0) needs to fly to L2-0. The truck in C0 is T0 and it is in L0-1 so it also needs to go to L1-0 to be loaded with P1.

5. Load-truck P1 to T2 at L2-1.
6. Drive-truck T2 from L2-1 to L2-0 in C2
7. Unload-truck P1 from T2 at L2-0
8. Fly-airplane A0 L1-0 to L2-0
9. Load-airplane P1 to A0 at L2-0
10. Fly-airplane A0 from L2-0 to L0-0
11. Unload-airplane P1 from A0 at L0-0
12. Drive-truck T0 from L0-1 to L0-0 in C0
13. Load-truck P1 to T0 at L0-0
14. Drive-truck T0 from L0-0 to L0-1 in C0
15. Unload-truck P1 from T0 at L0-1

Airplanes: A0 at L0-0, A1 at L1-0
Trucks: T0 at (driven but no change) L0-1, T1 at L1-0, T2 at L2-1
Packages: P3 at L0-0, P4 at L2-0, P5 at L2-0

Now, let's continue with P3. P3 needs to go from L0-0 to L2-0. The airplane A0 is already at L0-0, so we can directly load it to take it to L2-0 which is the final destination.

16. Load-airplane P3 to A0 at L0-0
17. Fly-airplane A0 from L0-0 to L2-0
18. Unload-airplane P3 from A0 at L2-0

Airplanes: A0 at L2-0, A1 at L1-0
Trucks: T0 at L0-1, T1 at L1-0, T2 at L2-1
Packages: P4 at L2-0, P5 at L2-0

Now, let's continue with P4. P4 needs to go from L2-0 to L1-0. The airplane A0 is already at L2-0, so we can directly load it to take it to L1-0 which is the final destination.

19. Load-airplane P4 to A0 at L2-0
20. Fly-airplane A0 from L2-0 to L1-0
21. Unload-airplane P4 from A0 at L1-0

Airplanes: A0 at L1-0, A1 at L1-0
Trucks: T0 at L0-1, T1 at L1-0, T2 at L2-1
Packages: P5 at L2-0

Finally, let's move P5. P5 needs to go from L2-0 to L1-1. The airplane A0 is at L1-0, so we need to fly it to L2-0 to load P5. Then, we need to fly it back to L1-0 and use the truck T1 to take it to L1-1. The truck in C1 is at L1-0, so it can directly br used to take P5 to the final destination L1-1.

22. Fly-airplane A0 from L1-0 to L2-0
23. Load-airplane P5 to A0 at L2-0
24. Fly-airplane A0 from L2-0 to L1-0
25. Unload-airplane P5 from A0 at L1-0
26. Load-truck P5 to T1 at L1-0
27. Drive-truck T1 from L1-0 to L1-1 in C1
28. Unload-truck P5 from T1 at L1-1

Airplanes: A0 at L1-0, A1 at L1-0
Trucks: T0 at L0-1, T1 at L1-1, T2 at L2-1
Packages: All packages are at their goal locations.

The PDDL version:
(drive-truck T1 L1-0 L1-1 C1)
(load-truck P2 T1 L1-1)
(drive-truck T1 L1-1 L1-0 C1)
(unload-truck P2 T1 L1-0)
(load-truck P1 T2 L2-1)
(drive-truck T2 L2-1 L2-0 C2)
(unload-truck P1 T2 L2-0)
(fly-airplane A0 L1-0 L2-0)
(load-airplane P1 A0 L2-0)
(fly-airplane A0 L2-0 L0-0)
(unload-airplane P1 A0 L0-0)
(drive-truck T0 L0-1 L0-0 C0)
(load-truck P1 T0 L0-0)
(drive-truck T0 L0-0 L0-1 C0)
(unload-truck P1 T0 L0-1)
(load-airplane P3 A0 L0-0)
(fly-airplane A0 L0-0 L2-0)
(unload-airplane P3 A0 L2-0)
(load-airplane P4 A0 L2-0)
(fly-airplane A0 L2-0 L1-0)
(unload-airplane P4 A0 L1-0)
(fly-airplane A0 L1-0 L2-0)
(load-airplane P5 A0 L2-0)
(fly-airplane A0 L2-0 L1-0)
(unload-airplane P5 A0 L1-0)
(load-truck P5 T1 L1-0)
(drive-truck T1 L1-0 L1-1 C1)
(unload-truck P5 T1 L1-1)
\end{assistantmessage}

\begin{usermessage}
Start:
Airplanes: A0 at L2-0, A1 at L2-0
Trucks: T0 at L0-0, T1 at L1-0, T2 L2-0
Packages: P0 at L0-0, P1 at L2-0, P2 at L2-0, P3 at L1-2, P4 at L0-0

Goal: P0 at L0-0, P1 at L0-0, P2 at L2-0, P3 at L2-1, P4 at L2-2
\end{usermessage}

\begin{assistantmessage}
5. Load-truck P3 to T1 at L1-2
6. Drive-truck T1 from L1-2 to L1-0 in C1
7. Unload-truck P3 from T1 at L1-0
8. Fly-airplane A1 from L2-0 to L1-0
9. Load-airplane P3 to A1 at L1-0
10. Fly-airplane A1 from L1-0 to L2-0
11. Unload-airplane P3 from A1 at L2-0
12. Load-truck P3 to T2 at L2-0
13. Drive-truck T2 from L2-0 to L2-1 in C2
14. Unload-truck P3 from T2 at L2-1

Airplanes: A0 at L0-0, A1 at L2-0
Trucks: T0 at L0-0, T1 at L1-0, T2 at L2-1
Packages: P4 at L0-0

Finally, let's move P4. P4 needs to go from L0-0 to L2-2. The airplane A0 is already at L0-0, so we can directly load it to take it to L2-0. Then, we need to use the truck T2 to take it from L2-0 to L2-2. The truck in C2 is at L2-1, so it needs to go to L2-0 to be loaded with P4 to take it to the final destination L2-2.

15. Load-airplane P4 to A0 at L0-0
16. Fly-airplane A0 from L0-0 to L2-0
17. Unload-airplane P4 from A0 at L2-0
18. Drive-truck T2 from L2-1 to L2-0 in C2
19. Load-truck P4 to T2 at L2-0
20. Drive-truck T2 from L2-0 to L2-2 in C2
21. Unload-truck P4 from T2 at L2-2

Airplanes: A0 at L2-0, A1 at L2-0
Trucks: T0 at L0-0, T1 at L1-0, T2 at L2-2
Packages: All packages are at their goal locations.

The PDDL version:
(load-airplane P1 A0 L2-0)
(fly-airplane A0 L2-0 L0-0)
(unload-airplane P1 A0 L0-0)
(drive-truck T1 L1-0 L1-2 C1)
(load-truck P3 T1 L1-2)
(drive-truck T1 L1-2 L1-0 C1)
(unload-truck P3 T1 L1-0)
(fly-airplane A1 L2-0 L1-0)
(load-airplane P3 A1 L1-0)
(fly-airplane A1 L1-0 L2-0)
(unload-airplane P3 A1 L2-0)
(load-truck P3 T2 L2-0)
(drive-truck T2 L2-0 L2-1 C2)
(unload-truck P3 T2 L2-1)
(load-airplane P4 A0 L0-0)
(fly-airplane A0 L0-0 L2-0)
(unload-airplane P4 A0 L2-0)
(drive-truck T2 L2-1 L2-0 C2)
(load-truck P4 T2 L2-0)
(drive-truck T2 L2-0 L2-2 C2)
(unload-truck P4 T2 L2-2)
\end{assistantmessage}

\begin{usermessage}
Start:
Airplanes: A0 at L1-0
Trucks: T0 at L0-0, T1 at L1-1
Packages: P0 at L1-0, P1 at L1-1, P2 at L0-0, P3 at L0-1

Goal: P0 at L0-1, P1 at L0-1, P2 at L1-0, P3 at L1-1
\end{usermessage}

\begin{assistantmessage}
There are two cities since only locations are talked about are L0 and L1. We have one airplane A0, and it is in L1-0. Let's first examine which packages need to moved to other cities.

P0: L1 -> L0 (yes)
P1: L1 -> L0 (yes)
P2: L0 -> L1 (yes)
P3: L0 -> L1 (yes)

All packages need to be transferred via planes to other cities. Let's transfer P0 first. P0 needs to go from L1-0 to L0-1. The airplane A0 is already at L1-0, so we can directly load it to take it to L0-0. Then, we need to use the truck T0 to take it from L0-0 to L0-1. The truck in C0 is T0 and it's at L0-0. So, the package P0 can be directly loaded to T0 to be taken to the final destination L0-1.

1. Load-airplane P0 to A0 at L1-0
2. Fly-airplane A0 from L1-0 to L0-0
3. Unload-airplane P0 from A0 at L0-0
4. Load-truck P0 to T0 at L0-0
5. Drive-truck T0 from L0-0 to L0-1 in C0
6. Unload-truck P0 from T0 at L0-1

Airplanes: A0 at L0-0
Trucks: T0 at L0-1, T1 at L1-1
Packages: P1 at L1-1, P2 at L0-0, P3 at L0-1

Now, let's continue with P1. P1 needs to go from L1-1 to L0-1. The truck in C1 is T1 and it's in L1-1 as the package, so we can directly load the package to T1. The airplane A0 is at L0-0, so we need to fly it to L1-0 to load P1. Then, we need to fly it back to L0-0 and use the truck T0 to take it to L0-1. The truck in C0 is T0 and it's in L0-1. So, it needs to be moved from L0-1 to L0-0 to load P1 and to be driven back to L0-1 which is the target destination for P1.

7. Load-truck P1 to T1 at L1-1
8. Drive-truck T1 from L1-1 to L1-0 in C1
9. Unload-truck P1 from T1 at L1-0
10. Fly-airplane A0 from L0-0 to L1-0
11. Load-airplane P1 to A0 at L1-0
12. Fly-airplane A0 from L1-0 to L0-0
13. Unload-airplane P1 from A0 at L0-0
14. Drive-truck T0 from L0-1 to L0-0 in C0
15. Load-truck P1 to T0 at L0-0
16. Drive-truck T0 from L0-0 to L0-1 in C0
17. Unload-truck P1 from T0 at L0-1

Airplanes: A0 at L0-0
Trucks: T0 at L0-1, T1 at L1-0
Packages: P2 at L0-0, P3 at L0-1

Now, let's continue with P2. P2 needs to go from L0-0 to L1-0. The airplane A0 is already at L0-0, so we can directly load it to take it to L1-0 which is the final destination.

18. Load-airplane P2 to A0 at L0-0
19. Fly-airplane A0 from L0-0 to L1-0
20. Unload-airplane P2 from A0 at L1-0

Airplanes: A0 at L1-0
Trucks: T0 at L0-1, T1 at L1-0
Packages: P3 at L0-1

Finally, let's move P3. P3 needs to go from L0-1 to L1-1. The truck in C0 is T0 and it's in L0-1 already. So, we can  directly load P3 to T0 to transfer it to L0-0. The airplane A0 is in L1-0, so it needs to be transferred to L0-0 to load P3 to transfer it to L1-0. The truck in C1 is T1 and it's already in L1-0, so we can directly load P3 to it to carry it to the target destination L1-1.

21. Load-truck P3 to T0 at L0-1
22. Drive-truck T0 from L0-1 to L0-0 in C0
23. Unload-truck P3 from T0 at L0-0
24. Fly-airplane A0 from L1-0 to L0-0
25. Load-airplane P3 to A0 at L0-0
26. Fly-airplane A0 from L0-0 to L1-0
27. Unload-airplane P3 from A0 at L1-0
28. Load-truck P3 to T1 at L1-0
29. Drive-truck T1 from L1-0 to L1-1 in C1
30. Unload-truck P3 from T1 at L1-1

Airplanes: A0 at L1-0
Trucks: T0 at L0-0, T1 at L1-1
Packages: All packages are at their goal locations.

The PDDL version:
(load-airplane P0 A0 L1-0)
(fly-airplane A0 L1-0 L0-0)
(unload-airplane P0 A0 L0-0)
(load-truck P0 T0 L0-0)
(drive-truck T0 L0-0 L0-1 C0)
(unload-truck P0 T0 L0-1)
(load-truck P1 T1 L1-1)
(drive-truck T1 L1-1 L1-0 C1)
(unload-truck P1 T1 L1-0)
(fly-airplane A0 L0-0 L1-0)
(load-airplane P1 A0 L1-0)
(fly-airplane A0 L1-0 L0-0)
(unload-airplane P1 A0 L0-0)
(drive-truck T0 L0-1 L0-0 C0)
(load-truck P1 T0 L0-0)
(drive-truck T0 L0-0 L0-1 C0)
(unload-truck P1 T0 L0-1)
(load-airplane P2 A0 L0-0)
(fly-airplane A0 L0-0 L1-0)
(unload-airplane P2 A0 L1-0)
(load-truck P3 T0 L0-1)
(drive-truck T0 L0-1 L0-0 C0)
(unload-truck P3 T0 L0-0)
(fly-airplane A0 L1-0 L0-0)
(load-airplane P3 A0 L0-0)
(fly-airplane A0 L0-0 L1-0)
(unload-airplane P3 A0 L1-0)
(load-truck P3 T1 L1-0)
(drive-truck T1 L1-0 L1-1 C1)
(unload-truck P3 T1 L1-1)
\end{assistantmessage}

\subsection{AoT+ Acre}

\begin{systemmessage}
Here, you will be given combinations of objects and whether the system will be on the "on" or "off" state. There are some set of objects that will lead to the system being in the "on" state. And if any of those objects are in the mix with the other objects, the result will always be "on". If none of them are there, then it's in the "off" state.

While solving, there will also be undetermined cases for some blocks. Here is the general idea:

If any one of the blocks is "on", the output is "on"
If all blocks are off, the output is "off"
If there is an (undetermined) block with all other "off" blocks, the output is "undetermined"

be careful, sometimes there are cases where we can understand some union of the undetermined blocks will be on. Follow the output exactly as it is given in the examples after writing "Test cases:
". I dont want you to write it in any other way.
\end{systemmessage}

\begin{usermessage}
Train:
blue rubber sphere, brown rubber cylinder -> on
brown rubber cylinder -> off
blue rubber sphere -> on
yellow metal sphere, green metal sphere, green rubber cylinder, cyan rubber cylinder, green rubber sphere -> on
green rubber sphere, brown metal cylinder, green rubber cylinder, cyan rubber cylinder -> off
cyan rubber cylinder, green rubber cylinder -> off

Test:
green rubber sphere -> ?
green metal sphere -> ?
brown rubber cylinder, green rubber cylinder -> ?
brown rubber cylinder, green rubber sphere -> ?
\end{usermessage}

\begin{assistantmessage}
Let's start with writing all the elements we know lead to the off or on state:
Off: brown rubber cylinder, green rubber sphere, brown metal cylinder, green rubber cylinder, cyan rubber cylinder
On: blue rubber sphere
undetermined: yellow metal sphere, green metal sphere

Now, we can focus on the line:
blue rubber sphere (on), brown rubber cylinder (off) -> on: does not yield any new info

Now, let's focus on the line:
yellow metal sphere (undetermined), green metal sphere (undetermined), green rubber cylinder (off), cyan rubber cylinder (off), green rubber sphere (off) -> on: at least one of the "yellow metal sphere" and "green metal sphere" are on. Can be one of them or both of them. This means that if those two appear together, that case will be on. If only one of them appear with the off objects it will be undetermined.

Test cases:
green rubber sphere -> off
green metal sphere (undetermined) -> undetermined
brown rubber cylinder (off), green rubber cylinder (off) -> off
brown rubber cylinder (off), green rubber sphere (off) -> off
\end{assistantmessage}

\begin{usermessage}
Train:
purple metal cube -> off
green rubber cube -> on
green rubber cube, purple metal cube -> on
green metal cube, gray metal cylinder, green rubber cylinder -> off
gray metal cube, gray metal cylinder, green rubber cylinder -> on
cyan metal cylinder, green metal cube, gray metal cylinder -> off

Test:
purple metal cube -> ?
green rubber cube -> ?
green metal cube, gray metal cylinder, green rubber cylinder, cyan metal cylinder -> ?
green metal cube, gray metal cylinder, green rubber cylinder, gray metal cube -> ?
\end{usermessage}

\begin{assistantmessage}
Let's start with writing all the elements we know lead to the off or on state:
Off: purple metal cube, green metal cube, gray metal cylinder, green rubber cylinder, cyan metal cylinder
On: green rubber cube, gray metal cube
undetermined: none

Then we already know every case.

Test cases:
purple metal cube (off) -> off
green rubber cube (on) -> on
green metal cube (off), gray metal cylinder (off), green rubber cylinder (off), cyan metal cylinder (off) -> off
green metal cube (off), gray metal cylinder (off), green rubber cylinder (off), gray metal cube (on) -> on
\end{assistantmessage}

\begin{usermessage}
Train:
yellow metal cube -> on
yellow metal cube, green rubber sphere -> on
green rubber sphere -> off
green metal cube -> on
yellow rubber sphere, green metal cube, yellow rubber cube, yellow rubber cylinder -> on
blue metal cylinder, yellow rubber cube, yellow rubber cylinder -> off

Test:
yellow rubber sphere -> ?
green metal cube -> ?
blue metal cylinder, yellow rubber cube, yellow rubber cylinder, yellow rubber sphere -> ?
blue metal cylinder, yellow rubber cube, yellow rubber cylinder, yellow metal cube -> ?
\end{usermessage}

\begin{assistantmessage}
Let's start with writing all the elements we know lead to the off or on state:
Off: green rubber sphere, blue metal cylinder, yellow rubber cube, yellow rubber cylinder
On: yellow metal cube, green metal cube
undetermined: yellow rubber sphere

Now, let's focus on the line:
yellow rubber sphere (undetermined), green metal cube (on), yellow rubber cube (off), yellow rubber cylinder (off) -> on: since one of them is already on, yellow rubber sphere stays undetermined

Test cases:
yellow rubber sphere (undetermined) -> undetermined
green metal cube (on) -> on
blue metal cylinder (off), yellow rubber cube (off), yellow rubber cylinder (off), yellow rubber sphere (undetermined) -> undetermined
blue metal cylinder (off), yellow rubber cube (off), yellow rubber cylinder (off), yellow metal cube (on) -> on
\end{assistantmessage}

\subsection{AoT+ List Functions}

\begin{systemmessage}
In the following, we will have lists as inputs and outputs. The task is to understand the rule to go from the input list to the output list. Tell the hypothesis that works for all of the input-output relations. In the end, always print out the python code as in the examples, do not change the style you give the python code.
\end{systemmessage}

\begin{usermessage}
[3, 4, 1, 5, 2, 0, 8, 6, 9] -> [1]
[5, 0, 6, 8, 2, 9, 4, 7, 3] -> [6]
[6, 3, 1, 4, 9, 0, 7] -> [1]
[8, 3, 9, 1, 7, 0] -> [9]
[9, 5, 7, 2] -> [7]
[6, 4, 9, 0, 5, 7, 3, 2, 8, 1] -> [9]
[8, 0, 6, 4, 7, 5, 3, 9, 2] -> [6]
[8, 6, 1, 0, 5, 9] -> [1]
\end{usermessage}

\begin{assistantmessage}
First obvious thing is that although the input length varies, the output always has one element. Let's first see whether the output element is included in the input and whether the place it appears has some relation:

[3, 4, 1, 5, 2, 0, 8, 6, 9] -> [1] appears in the third place
[5, 0, 6, 8, 2, 9, 4, 7, 3] -> [6] appears in the third place
[6, 3, 1, 4, 9, 0, 7] -> [1] appears in the third place
[8, 3, 9, 1, 7, 0] -> [9] appears in the third place
[9, 5, 7, 2] -> [7] appears in the third place
[6, 4, 9, 0, 5, 7, 3, 2, 8, 1] -> [9] appears in the third place
[8, 0, 6, 4, 7, 5, 3, 9, 2] -> [6] appears in the third place
[8, 6, 1, 0, 5, 9] -> [1] appears in the third place

Since it always appear in the third place, that's our hypothesis. However, there are no examples as to what happens when there are less than three elements in the input. By default, let's say we will return an empty list.

Final hypothesis: If the input list has at least three elements, than the output is the third element. If it has less than three elements, than the output is an empty list.

Python program:
\begin{lstlisting}
def transform_list(input_lst):
    if len(input_lst) >= 3:
        return [input_lst[2]]
    else:
        return []
\end{lstlisting}
\end{assistantmessage}

\begin{usermessage}
[7, 0, 3, 8, 9, 2, 6, 5] -> [7, 0, 3, 8, 9, 2, 4, 6, 5]
[0] -> [0, 4]
[8, 2, 5, 9, 3, 4, 6, 0, 7] -> [8, 2, 5, 9, 3, 4, 4, 6, 0, 7]
[2, 5] -> [2, 5, 4]
[0, 8, 3, 5, 2, 9, 1, 7, 6, 4] -> [0, 8, 3, 5, 2, 9, 4, 1, 7, 6, 4]
[7, 5, 2, 6, 0, 3, 9, 1, 8, 4] -> [7, 5, 2, 6, 0, 3, 4, 9, 1, 8, 4]
[5, 1, 9] -> [5, 1, 9, 4]
[] -> [4]
\end{usermessage}

\begin{assistantmessage}
The input and the output has variable length. Let's first analyze the input and output lengths:
8 -> 9
1 -> 2
8 -> 9
2 -> 3
10 -> 11
10 -> 11
3 -> 4
0 -> 1

In all cases, the output list is one element longer than the input list. Let's see where the new element is added and what it is:

[7, 0, 3, 8, 9, 2, 6, 5] -> [7, 0, 3, 8, 9, 2, 4, 6, 5] 4 is added at the 7th place
[0] -> [0, 4] 4 is added at the 2nd place
[8, 2, 5, 9, 3, 4, 6, 0, 7] -> [8, 2, 5, 9, 3, 4, 4, 6, 0, 7] 4 is added at the 7th place
[2, 5] -> [2, 5, 4] 4 is added at the 3rd place
[0, 8, 3, 5, 2, 9, 1, 7, 6, 4] -> [0, 8, 3, 5, 2, 9, 4, 1, 7, 6, 4] 4 is added at the 7th place
[7, 5, 2, 6, 0, 3, 9, 1, 8, 4] -> [7, 5, 2, 6, 0, 3, 4, 9, 1, 8, 4] 4 is added at the 7th place
[5, 1, 9] -> [5, 1, 9, 4] 4 is added at the 4th place
[] -> [4] 4 is added at the 1st place

The added element is always 4. It is added at the 7th place if the list has at least 7 elements, otherwise it is added at the end of the list.

Final hypothesis: The output list is the input list with 4 added at the 7th place if the list has at least 7 elements, otherwise 4 is added at the end of the list.

Python program:
\begin{lstlisting}
def transform_list(input_lst):
    if len(input_lst) >= 7:
        return input_lst[:6] + [4] + input_lst[6:]
    else:
        return input_lst + [4]
\end{lstlisting}
\end{assistantmessage}

\begin{usermessage}
[42, 1, 42, 42, 18, 42, 1] -> [1, 42, 42, 18, 42, 1, 42]
[4, 0] -> [4, 0]
[65, 0, 8, 59, 7, 1] -> [0, 8, 59, 7, 1, 65]
[9, 0, 2] -> [2, 9, 0]
[7, 9, 63, 1] -> [9, 63, 1, 7]
[34, 68, 1, 4, 0] -> [34, 68, 1, 4, 0]
[75, 0, 45, 5, 3, 2] -> [45, 5, 3, 2, 75, 0]
[8, 9, 1] -> [9, 1, 8]
\end{usermessage}

\begin{assistantmessage}
The input and the output has variable length. Let's first analyze the input and output lengths:
7 -> 7
2 -> 2
6 -> 6
3 -> 3
4 -> 4
5 -> 5
6 -> 6
3 -> 3

In all cases, the output list is the same length as the input list. Let's see whether the elements in the output list are included in the input list and whether the order has changed:

[42, 1, 42, 42, 18, 42, 1] -> [1, 42, 42, 18, 42, 1, 42] There is definitely many similarities between the input and the output. The input is shifted to the left by 1
[4, 0] -> [4, 0] They are the same
[65, 0, 8, 59, 7, 1] -> [0, 8, 59, 7, 1, 65] The input is shifted to the left by 1
[9, 0, 2] -> [2, 9, 0] The input is shifted to the left by 2
[7, 9, 63, 1] -> [9, 63, 1, 7] The input is shifted to the left by 1
[34, 68, 1, 4, 0] -> [34, 68, 1, 4, 0] They are the same
[75, 0, 45, 5, 3, 2] -> [45, 5, 3, 2, 75, 0] The input is shifted to the left by 2
[8, 9, 1] -> [9, 1, 8] The input is shifted to the left by 1

The input is either shifted to the left by some amount or stays the same (or we can think of it as shifted to the left by 0. Let's see whether there is some pattern for how much it is shifted. Perhaps that shifting amount appears in the input sequence.

[42, 1, 42, 42, 18, 42, 1] -> shifted by 1 -> the second and the last element (7th element)
[4, 0] -> shifted by 0 -> the second or the last element
[65, 0, 8, 59, 7, 1] -> shifted by 1 -> 6th or the last element
[9, 0, 2] -> shifted by 2 -> third or the last element
[7, 9, 63, 1] -> shifted by 1 -> fourth or the last element
[34, 68, 1, 4, 0] -> shifted by 0 -> fifth or the last element
[75, 0, 45, 5, 3, 2] -> shifted by 2 -> 6th or the last element
[8, 9, 1] -> [9, 1, 8] -> shifted by 1 -> third or the last element

The shifting amount is always the last element in the input list. 

Final hypothesis: The output list is the input list shifted to the left by the amount that is the last element in the input list.

Python program:
\begin{lstlisting}
def transform_list(input_lst):
    if len(input_lst) == 0:
        return []
    shift_amount = input_lst[-1]
    return input_lst[shift_amount:] + input_lst[:shift_amount]
\end{lstlisting}
\end{assistantmessage}

\begin{usermessage}
[6, 5, 11, 42, 91, 70, 4] -> [11, 42, 91]
[3, 91, 25, 90, 92, 0] -> [25, 90]
[6, 74, 55, 80, 2, 5, 60] -> [55, 80, 2]
[22, 21, 62, 6, 0] -> [62]
[5, 24, 78, 39, 8] -> [78]
[24, 46, 0, 31, 74, 5, 44, 80, 94] -> [0, 31, 74, 5, 44]
[93, 23, 36, 63, 40, 25, 9, 43, 8, 53] -> [36, 63, 40, 25, 9, 43]
[3, 35, 10, 17, 12, 0, 46, 6, 8, 7] -> [10, 17, 12, 0, 46, 6]
\end{usermessage}

\begin{assistantmessage}
The input and the output has variable length. Let's first analyze the input and output lengths:
7 -> 3
6 -> 2
7 -> 3
5 -> 1
5 -> 1
9 -> 5
10 -> 6
10 -> 6

The output list is always shorter than the input list. Let's see whether the elements in the output list are included in the input list and whether the order has changed:

[6, 5, 11, 42, 91, 70, 4] -> [11, 42, 91] The output list is the input list without the first two and the last two elements
[3, 91, 25, 90, 92, 0] -> [25, 90] The output list is the input list without the first two and the last two elements
[6, 74, 55, 80, 2, 5, 60] -> [55, 80, 2] The output list is the input list without the first two and the last two elements
[22, 21, 62, 6, 0] -> [62] The output list is the input list without the first two and the last two elements
[5, 24, 78, 39, 8] -> [78] The output list is the input list without the first two and the last two elements
[24, 46, 0, 31, 74, 5, 44, 80, 94] -> [0, 31, 74, 5, 44] The output list is the input list without the first two and the last two elements
[93, 23, 36, 63, 40, 25, 9, 43, 8, 53] -> [36, 63, 40, 25, 9, 43] The output list is the input list without the first two and the last two elements
[3, 35, 10, 17, 12, 0, 46, 6, 8, 7] -> [10, 17, 12, 0, 46, 6] The output list is the input list without the first two and the last two elements

The output list is always the input list without the first two and the last two elements. We should also consider what happens when the input has length less than 4. By default, let's return an empty list.
Final hypothesis: The output list is the input list without the first two and the last two elements. If the input list has less than 4 elements, the output is an empty list.

Python program:
\begin{lstlisting}
def transform_list(input_lst):
    if len(input_lst) < 4:
        return []
    else:
        return input_lst[2:-2]
\end{lstlisting}
\end{assistantmessage}

\begin{usermessage}
[15, 2, 1, 2, 2, 15, 6, 15, 4, 2] -> [1, 6, 4]
[1, 1, 1, 1, 1, 1, 1] -> []
[2, 2, 12, 3, 0, 5, 1, 4, 7] -> [12, 3, 0, 5, 1, 4, 7]
[9, 47, 7, 5, 47, 25, 5, 45, 25] -> [9, 7, 45]
[0, 8, 4, 2, 8, 4, 6] -> [0, 2, 6]
[18, 5, 5, 5, 5, 5, 18, 18, 5, 5] -> []
[8, 5, 8, 9, 8, 8] -> [5, 9]
[1, 95, 4, 0, 54, 4, 3, 54, 95, 95] -> [1, 0, 3]
\end{usermessage}

\begin{assistantmessage}
The input and the output has variable length. Let's first analyze the input and output lengths:
10 -> 3
7 -> 0
9 -> 7
9 -> 3
7 -> 3
10 -> 0
6 -> 2
10 -> 3

The output list is always shorter than the input list. Let's see whether the elements in the output list are included in the input list and whether the order has changed:

[15, 2, 1, 2, 2, 15, 6, 15, 4, 2] -> [1, 6, 4] The output list is the input list without the elements that appear more than once
[1, 1, 1, 1, 1, 1, 1] -> [] The output list is the input list without the elements that appear more than once
[2, 2, 12, 3, 0, 5, 1, 4, 7] -> [12, 3, 0, 5, 1, 4, 7] The output list is the input list without the elements that appear more than once
[9, 47, 7, 5, 47, 25, 5, 45, 25] -> [9, 7, 45] The output list is the input list without the elements that appear more than once
[0, 8, 4, 2, 8, 4, 6] -> [0, 2, 6] The output list is the input list without the elements that appear more than once
[18, 5, 5, 5, 5, 5, 18, 18, 5, 5] -> [] The output list is the input list without the elements that appear more than once
[8, 5, 8, 9, 8, 8] -> [5, 9] The output list is the input list without the elements that appear more than once
[1, 95, 4, 0, 54, 4, 3, 54, 95, 95] -> [1, 0, 3] The output list is the input list without the elements that appear more than once

The output list is always the input list without the elements that appear more than once. 

Final hypothesis: The output list is the input list without the elements that appear more than once.

Python program:
\begin{lstlisting}
def transform_list(input_lst):
    return [x for x in input_lst if input_lst.count(x) == 1]
\end{lstlisting}
\end{assistantmessage}

\begin{usermessage}
[2, 5, 3, 9] -> [3]
[2, 0, 6, 1, 9] -> [6]
[8, 9, 1, 4, 3, 2, 5, 7, 0, 6] -> [0]
[6, 8, 9, 3, 8, 7, 4, 1, 2] -> [4]
[3, 8, 2, 1, 7, 5, 6, 4] -> [1]
[2, 5, 3, 1, 0, 4, 7, 8] -> [3]
[1, 7, 3, 2, 5, 6, 8, 9, 0, 4] -> [7]
[4, 9, 2, 3, 6, 0] -> [6]
\end{usermessage}

\begin{assistantmessage}
First obvious thing is that although the input length varies, the output always has one element. Let's first see whether the output element is included in the input and whether the place it appears has some relation:

[2, 5, 3, 9] -> [3] appears in the third place
[2, 0, 6, 1, 9] -> [6] appears in the third place
[8, 9, 1, 4, 3, 2, 5, 7, 0, 6] -> [0] appears in the seventh place
[6, 8, 9, 3, 8, 7, 4, 1, 2] -> [4] appears in the seventh place
[3, 8, 2, 1, 7, 5, 6, 4] -> [1] appears in the fourth place
[2, 5, 3, 1, 0, 4, 7, 8] -> [3] appears in the third place
[1, 7, 3, 2, 5, 6, 8, 9, 0, 4] -> [7] appears in the second place
[4, 9, 2, 3, 6, 0] -> [6] appears in the fifth place

Since it doesn't always appear in the same place, let's see whether the place it appears has some relation with the first element in the input list:

[2, 5, 3, 9] -> [3] appears in the third place -> 2 + 1
[2, 0, 6, 1, 9] -> [6] appears in the third place -> 2 + 1
[8, 9, 1, 4, 3, 2, 5, 7, 0, 6] -> [0] appears in the seventh place -> 8 - 1
[6, 8, 9, 3, 8, 7, 4, 1, 2] -> [4] appears in the seventh place -> 6 + 1
[3, 8, 2, 1, 7, 5, 6, 4] -> [1] appears in the fourth place -> 3 + 1
[2, 5, 3, 1, 0, 4, 7, 8] -> [3] appears in the third place -> 2 + 1
[1, 7, 3, 2, 5, 6, 8, 9, 0, 4] -> [7] appears in the second place -> 1 + 1
[4, 9, 2, 3, 6, 0] -> [6] appears in the fifth place -> 4 + 1

The output element is always the element in the input list that is at the position of the first element plus one. However, there are no examples as to what happens when there are less elements in the input than the first element plus one. By default, let's say we will return an empty list.

Final hypothesis: If the input list has more elements than the first element plus one, than the output is the element at the position of the first element plus one (in python, it will directly be the element at the position of the first element). If it has less elements than the first element plus one, than the output is an empty list.

Important reminder: since we are talking about the order, let's not forget that python indexing starts from 0. Therefore, let's say 5th element should be accessed by 4, not 5.

Python program:
\begin{lstlisting}
def transform_list(input_lst):
    if len(input_lst) > input_lst[0]:
        return [input_lst[input_lst[0]]]
    else:
        return []
\end{lstlisting}
\end{assistantmessage}

\begin{usermessage}
[1, 9, 4, 7, 6, 3] -> [9, 4, 7]
[0, 2, 9, 6, 4, 8, 5, 1] -> [2, 9, 6]
[7, 3, 5, 0, 2] -> [3, 5, 0]
[7, 5, 4] -> [5, 4]
[7, 9, 3, 8, 4] -> [9, 3, 8]
[7, 6, 3, 8, 5, 9, 0, 4, 1] -> [6, 3, 8]
[6] -> []
[3, 9, 1, 7, 5, 4, 2] -> [9, 1, 7]
\end{usermessage}

\begin{assistantmessage}
The input and the output has variable length. Let's first analyze the input and output lengths:
6 -> 3
8 -> 3
5 -> 3
3 -> 2
5 -> 3
9 -> 3
1 -> 0
7 -> 3

The output list is always shorter than the input list. Let's see whether the elements in the output list are included in the input list and whether the order has changed:

[1, 9, 4, 7, 6, 3] -> [9, 4, 7] The output list is the input list without the first and the last two elements
[0, 2, 9, 6, 4, 8, 5, 1] -> [2, 9, 6] The output list is the input list without the first and the last four elements. Or They are at the positions 2, 3, 4
[7, 3, 5, 0, 2] -> [3, 5, 0] The output numbers at the positions 2, 3, 4
[7, 5, 4] -> [5, 4] The output numbers at the positions 2, 3
[7, 9, 3, 8, 4] -> [9, 3, 8] The output numbers at the positions 2, 3, 4
[7, 6, 3, 8, 5, 9, 0, 4, 1] -> [6, 3, 8] The output numbers at the positions 2, 3, 4
[6] -> [] The output is empty, because the input only has 1 element. 
[3, 9, 1, 7, 5, 4, 2] -> [9, 1, 7] The output numbers at the positions 2, 3, 4

Let's check whether the first one is also the same:
[1, 9, 4, 7, 6, 3] -> [9, 4, 7] The output numbers at the positions 2, 3, 4

So, the elements are at the positions 2, 3, 4 mostly, and once at 2, 3 (when the input only has 3 elements) and once the output is empty when the list only has a single element. It appears that the output consists of starting from the second position, but including at most three elements.

Final hypothesis: The output consists of elements that start from the second position of the input list, up to three elements at most.

Python program:
\begin{lstlisting}
def transform_list(input_lst):
    max_index = min(4, len(input_lst))
    if len(input_lst) >= 2:
        return input_lst[1:max_index]
    else:
        return []
\end{lstlisting}
\end{assistantmessage}

\end{document}